\documentclass[runningheads]{llncs}
\usepackage{graphicx}

\usepackage{tikz}
\usepackage{comment}
\usepackage{amsmath,amssymb} 
\usepackage{color}

\usepackage{stackengine}
\usepackage{multirow}
\usepackage{tabularx}
\usepackage{wrapfig}
\usepackage{cite}

\usepackage{makecell}

\usepackage{pifont}
\newcommand{\xmark}{\text{\ding{55}}}

\usepackage[makeroom]{cancel}

\begin{document}
\pagestyle{headings}
\mainmatter
\def\ECCVSubNumber{1738}  

\title{Learning Local Implicit Fourier Representation\\for Image Warping} 

\titlerunning{Local Texture Estimator for Image Warping}
%
\author{Jaewon Lee\inst{1}\and
Kwang Pyo Choi\inst{2}\and
Kyong Hwan Jin\inst{1}\thanks{Corresponding author.}
}
\authorrunning{J. Lee et al.}
%
\institute{{Daegu Gyeongbuk Institute of Science and Technology (DGIST), Korea \and
Samsung Electronics, Korea
}
\email{ljw3136@dgist.ac.kr,kp5.choi@samsung.com,kyong.jin@dgist.ac.kr}}
\maketitle

\vspace{-25pt}
\begin{abstract}
Image warping aims to reshape images defined on rectangular grids into arbitrary shapes. Recently, implicit neural functions have shown remarkable performances in representing images in a continuous manner. However, a standalone multi-layer perceptron suffers from learning high-frequency Fourier coefficients. In this paper, we propose a local texture estimator for image warping (LTEW) followed by an implicit neural representation to deform images into continuous shapes. Local textures estimated from a deep super-resolution (SR) backbone are multiplied by locally-varying Jacobian matrices of a coordinate transformation to predict Fourier responses of a warped image. Our LTEW-based neural function outperforms existing warping methods for asymmetric-scale SR and homography transform. Furthermore, our algorithm well generalizes arbitrary coordinate transformations, such as homography transform with a large magnification factor and equirectangular projection (ERP) perspective transform, which are not provided in training. Our source code is available at \textcolor{magenta}{\url{https://github.com/jaewon-lee-b/ltew}}.


\end{abstract}

\vspace{-25pt}
\section{Introduction}

Our goal is to deform images defined on rectangular grids into continuous shapes, referred to as image warping. Image warping is widely used in various computer vision and graphic tasks, such as image editing \cite{572000, CHIANG2000761}, optical flow \cite{Sun_2018_CVPR}, image alignment \cite{NIPS2015_33ceb07b, Sajjadi_2018_CVPR, Chan_2021_CVPR, Jiang_2021_ICCV}, and omnidirectional vision \cite{5652095, 5756235, NIPS2017_1113d7a7, Lee_2019_CVPR, Deng_2021_CVPR}. A conventional approach \cite{NIPS2015_33ceb07b, 1163711} applies an inverse coordinate transformation to interpolate the missing RGB value in the input space. However, interpolation-based methods cause jagging and blurring artifacts in output images. Recently, SRWarp \cite{SRWarp} paved the way to reshape images with high-frequency details by adopting a deep single image super-resolution (SISR) architecture as a backbone.

SISR is a particular case of image warping \cite{SRWarp}. 
A goal of SISR is to reconstruct a high-resolution (HR) image from its degraded low-resolution (LR) counterpart. Recent lines of research in solving SISR are to extract deep feature maps using advanced architectures \cite{Lim_2017_CVPR_Workshops, zhang2018residual,zhang2018rcan, Wang_2018_ECCV_Workshops, Mei_2021_CVPR, DBLP:conf/cvpr/Chen000DLMX0021, liang2021swinir} and upscale them to HR images at the end of a network \cite{DBLP:journals/corr/ShiCHTABRW16, hu2019meta, chen2021learning, lee2021local, itsrn}. Even though deep SISR methods reconstruct visually clear HR images, directly applying them to our problem is limited since each local region of a warped image is stretched with different scale factors \cite{4056910}. 
By reconsidering the warping problem as a spatially-varying SR task, SRWarp \cite{SRWarp} shed light on deforming images with sharp edges. However, interpolation-based SRWarp shows limited performance in generalizing to a large-scale representation which is out of training range.


Recently, implicit neural functions have attracted significant attention in representing signals, such as image \cite{chen2021learning, lee2021local}, video \cite{Nerv}, signed distance \cite{Park_2019_CVPR}, occupancy \cite{Occupancy_Networks}, shape \cite{Local_Implicit_Grid_CVPR20}, and view synthesis \cite{sitzmann2019srns, mildenhall2020nerf}, in a continuous manner. A multi-layer perceptron (MLP) parameterizes such an implicit neural representation \cite{sitzmann2019siren, tancik2020fourfeat} and takes coordinates as an input. Inspired by the recent implicit function success, LIIF \cite{chen2021learning} well generalizes to a large-scale rectangular SR beyond a training distribution.
However, one shortcoming of implicit neural functions \cite{DBLP:conf/icml/RahamanBADLHBC19, tancik2020fourfeat} is that a standalone MLP with ReLUs is biased towards learning low-frequency content.
To alleviate this spectral bias problem, Local Texture Estimator (LTE) \cite{lee2021local} estimates Fourier features for an HR image from its LR counterpart motivated by Fourier analysis. While LTE achieved arbitrary-scale rectangular SR with high-frequency details, LTE representation fails to evaluate a frequency response for image warping due to its spatially-varying nature.

\begin{figure}[t]
\centering
\includegraphics[scale = 0.177]{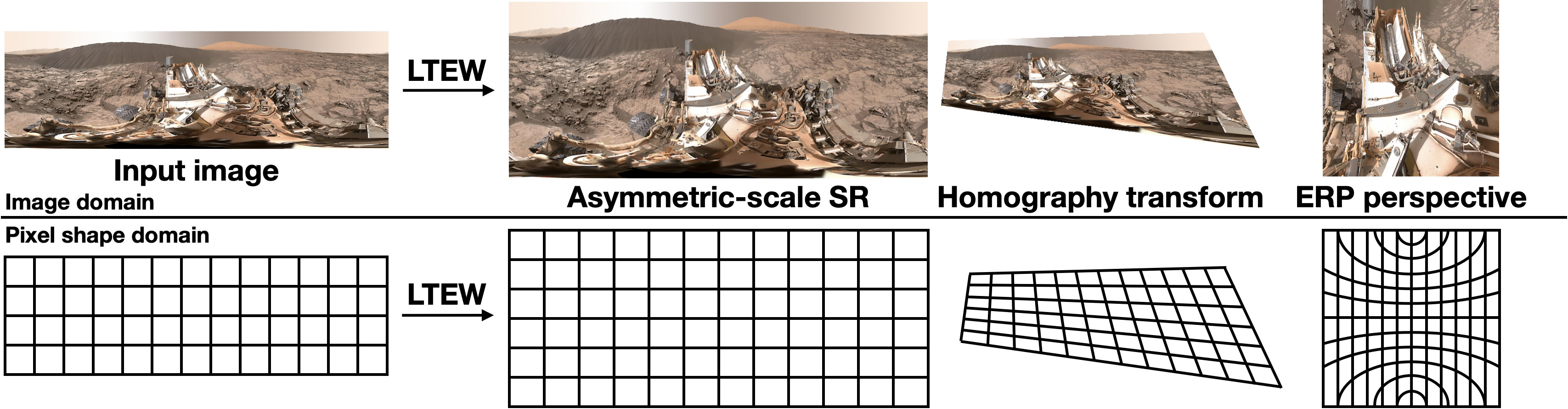}
\vspace{-6pt}
\caption{Implicit neural representation for image warping.}
\label{fig:overview}
\vspace{-12pt}
\end{figure}

Given an image $\mathbf{I^{IN}}:X\mapsto\mathbb{R}^3$ and a differentiable and invertible coordinate transformation $f:X\mapsto Y$, we propose a local texture estimator for image warping (LTEW) followed by an implicit neural function representing $\mathbf{I^{WARP}}:Y\mapsto\mathbb{R}^3$, as in Fig.~\ref{fig:overview}. Our algorithm leverages both Fourier features estimated from an input image and the Jacobian of coordinate transformation. In geometry, the determinant of the Jacobian indicates a local magnification ratio. Hence, we multiply spatially-varying Jacobian matrices to Fourier features for each pixel before our MLP represents $\mathbf{I^{WARP}}$. Furthermore, we point out that a spatially-varying prior for pixel shape is essential in enhancing a representational power of neural functions. The pixel shape described by orientation and curvature is numerically computed by gradient terms of given coordinate transformation.

We demonstrate that our proposed LTEW with a deep SISR backbone \cite{zhang2018residual, zhang2018rcan, Wang_2018_ECCV_Workshops} surpasses existing warping methods \cite{1163711, Wang2020Learning, SRWarp} for both upscaling and homography transform. While previous warping techniques \cite{Wang2020Learning, SRWarp} employ convolution and polynomial interpolation as a resampling module, our LTEW-based implicit neural function takes continuous coordinates as an input. Therefore, our proposed algorithm effectively generalizes in continuously representing $\mathbf{I^{WARP}}$, especially for homography transforms with a substantial magnification factor ($\times 4$-$\times 8$), which is not provided during a training phase. We further pay attention to omnidirectional imaging (ODI) \cite{5652095, 5756235, NIPS2017_1113d7a7, Lee_2019_CVPR, Deng_2021_CVPR} to verify the generalization ability of our algorithm. With the rapid advancement in virtual reality (VR), ODI has become crucial for product development. Equirectangular projection (ERP) is widely used in imaging pipelines of a head-mounted display (HMD) \cite{8434258}. As a result of projection from spherical grids to rectangular grids, pixels are sparsely located near high latitudes. Since the proposed LTEW learns spatially-varying properties, our method qualitatively outperforms other warping methods in projecting perspective without extra training.

To summarize, the contributions of our work include:
\begin{itemize}
    \item We propose a continuous neural representation for image warping by taking advantage of both Fourier features and spatially-varying Jacobian matrices of coordinate transformations.
    \item We claim that a spatially-varying prior for pixel shape described by orientation and curvature is significant in improving the representational capacity of the neural function.
    \item We demonstrate that our LTEW-based implicit neural function outperforms the existing warping methods for upscaling and homography transform, and unseen coordinate transformations.
\end{itemize}

\section{Related Works}
\textbf{Image warping} Image warping is a popular technique for various computer vision and graphics tasks, such as image editing \cite{572000, CHIANG2000761}, optical flow estimation \cite{Sun_2018_CVPR}, and image alignment \cite{NIPS2015_33ceb07b, Sajjadi_2018_CVPR, Chan_2021_CVPR, Jiang_2021_ICCV}. A general technique for image warping \cite{NIPS2015_33ceb07b} is finding a spatial location in input space and applying an interpolation kernel to calculate missing RGB values. Even though an interpolation-based image warping is differentiable and an easy-to-implement framework, an output image suffers from jagging and blurring artifacts \cite{SRWarp}. Recently, SRWarp \cite{SRWarp} proposed an arbitrary image transformation framework by interpreting an image warping task as a spatially-varying SR problem. SRWarp shows noticeable performance gain in arbitrary SR, including homography transform using an adaptive warping layer. However, the generalization ability of SRWarp is limited for unseen transformations, like a homography transform with a large magnification factor.

\vspace{10pt}
\noindent
\textbf{Implicit neural representation (INR)} Motivated from the fact that neural network is a universal function approximator \cite{10.5555/70405.70408}, INR is widely applied to represent continuous-domain signals. Conventionally, the memory requirement for data is quadratically (2D) or cubically (3D) proportional to signal resolution. In contrast, INR is a memory-efficient framework to store continuous signals since storage size is proportional to the number of model parameters rather than signal resolution. Recently, local INR \cite{Local_Implicit_Grid_CVPR20, chen2021learning, lee2021local} has been proposed to enhance the spatial resolution of input signals in an arbitrary manner. By using both feature maps from a deep neural encoder and relative coordinates (or \textit{local grid} in \cite{Local_Implicit_Grid_CVPR20, Lee_2019_CVPR}), such approaches are capable of generalizing to unseen tasks, which are not given during training. Inspired by previous works, our proposed LTEW utilizes Fourier features from a deep neural backbone and local grids to represent warped images under arbitrary coordinate transformations.

\vspace{10pt}
\noindent
\textbf{Spectral bias} Early works \cite{DBLP:conf/icml/RahamanBADLHBC19, tancik2020fourfeat} have shown that INR parameterized by a standalone MLP with a ReLU activation fails to capture high-frequency details of signals. Dominant approaches for resolving this spectral bias problem are substituting ReLUs with a periodic function \cite{sitzmann2019siren}, projecting input coordinates into a high-dimensional Fourier \cite{mildenhall2020nerf, tancik2020fourfeat, Benbarka_2022_WACV, lee2021local} or spline \cite{ijcai2021-151} feature space, and multiplying sinusoidal or Gabor filters \cite{fathony2021multiplicative}. Recently, LTE \cite{lee2021local} achieved arbitrary-scale SR using INR by estimating Fourier information from an LR image. Unlike previous attempts \cite{mildenhall2020nerf, tancik2020fourfeat}, Fourier feature space in LTE representation \cite{lee2021local} is data-driven and characterizes texture maps in 2D space. However, considering a spatially-varying SR issue in image warping, directly applying LTE is limited to characterize the Fourier space of warped images.

\vspace{10pt}
\noindent
\textbf{Deep SISR} After ESPCN \cite{DBLP:journals/corr/ShiCHTABRW16} has proposed a memory-efficient upsampling layer based on sub-pixel convolution, advanced deep vision backbones, such as residual block \cite{Lim_2017_CVPR_Workshops}, densely connected residual block \cite{zhang2018residual, Wang_2018_ECCV_Workshops}, channel attention \cite{zhang2018rcan}, second-order attention \cite{dai2019second}, holistic attention \cite{han2020}, non-local network \cite{Mei_2021_CVPR}, are jointly employed to reconstruct high-quality images. Recently, by taking advantage of inductive bias in self-attention mechanism \cite{DBLP:conf/iclr/DosovitskiyB0WZ21, liu2021Swin}, general-purpose image restoration backbones \cite{DBLP:conf/cvpr/Chen000DLMX0021, liang2021swinir} remarkably outperform convolution-based architectures using a large dataset. Despite their compelling representational power, we have to train and store several models for each scale factor. Current approaches \cite{hu2019meta, Wang2020Learning, chen2021learning, lee2021local, itsrn} for arbitrary-scale SR are utilizing a dynamic filter network \cite{hu2019meta, Wang2020Learning}, INR \cite{chen2021learning, lee2021local}, or transformer \cite{itsrn}. Unlike the previous arbitrary-scale SR methods, our LTEW represents images under arbitrary coordinate transformations, like homography transform, with only a single network.

\vspace{10pt}
\noindent
\textbf{Omnidirectional image (ODI)} In a new era of VR, omnidirectional vision \cite{5652095, 5756235, NIPS2017_1113d7a7, Lee_2019_CVPR, Deng_2021_CVPR} becomes playing a crucial role in product development. While natural images are represented in the Euclidean space, ODIs are defined on the spherical coordinates. A common methods to project ODIs to the 2D plane is an ERP to be consistent with imaging pipelines in HMD \cite{8434258}. One limitation of an ERP projected ODIs is that non-uniform spatial resolving power leads to severe spatial distortion near boundaries \cite{Lee_2019_CVPR}. To handle this varying pixel densities across latitudes, Deng \textit{et al.} \cite{Deng_2021_CVPR} proposed a hierarchically adaptive network. Since our proposed LTEW utilizes Jacobian matrices of given coordinate transformation to learn spatially-varying property for image warping, the distortion caused by varying pixel densities can be safely projected without extra training.

\section{Problem Formulation}
\label{sec:prob}
\begin{figure}[t]
\centering
\includegraphics[scale = 0.177]{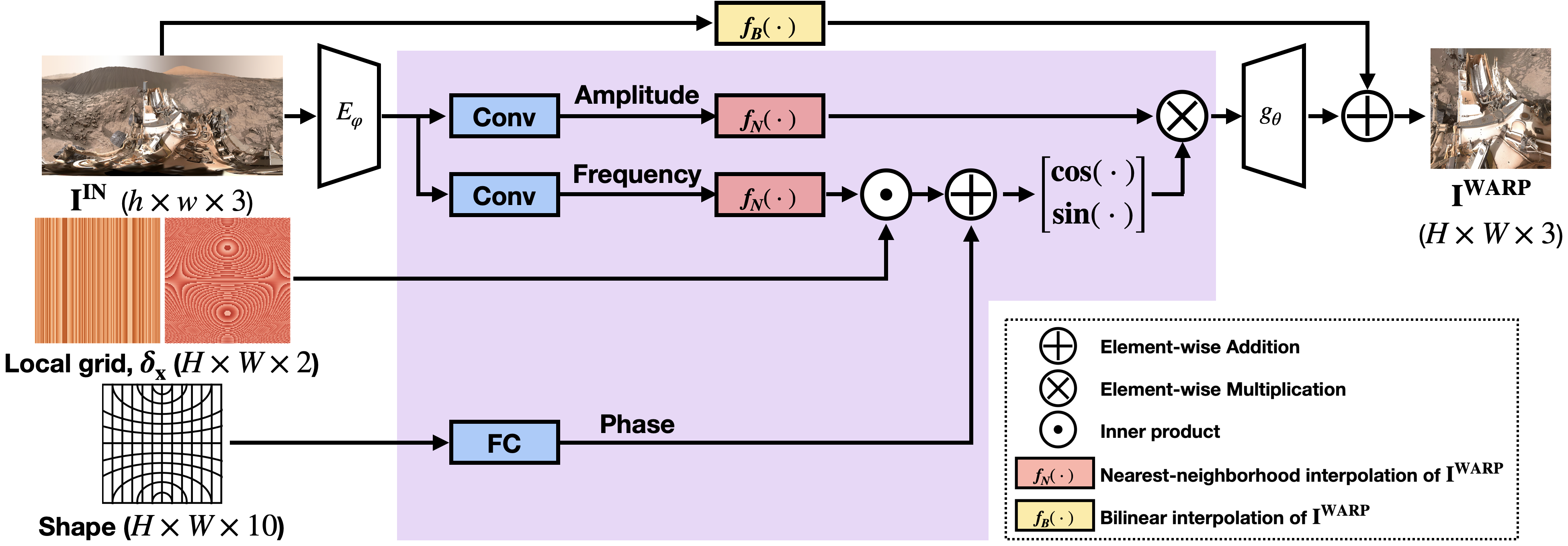}
\vspace{-15pt}
\caption{{A flowchart of our proposed Local Texture Estimator Warp (LTEW).}}
\label{fig:concept}
\end{figure}

Given an image $\mathbf {I^{IN}}:X\mapsto\mathbb{R}^3$, and a differentiable and invertible coordinate transformation $f:X\mapsto Y$, our goal is to formulate an implicit neural representation of an $\mathbf {I^{WARP}}:Y\mapsto\mathbb{R}^3$ for image warping.
The set $X\triangleq\{\mathbf x|\mathbf x\in\mathbb{R}^2 \}:[0, h)\times [0, w)$ is an input coordinate space, and $Y\triangleq\{\mathbf y|\mathbf y = f(\mathbf x)\in\mathbb{R}^2 \}:[0, H)\times [0, W)$ is an output coordinate space.
In practice, warping changes an image resolution to preserve the density of pixels (In Fig.~\ref{fig:concept}: $h\times w\rightarrow H\times W$).
Note that neural representation parameterized by an MLP with ReLU activations fails to capture high-frequency details of signals \cite{DBLP:conf/icml/RahamanBADLHBC19, mildenhall2020nerf, sitzmann2019siren, tancik2020fourfeat}. Recently, LTE \cite{lee2021local} achieved arbitrary-scale SR for symmetric scale factors by estimating essential Fourier information. However, concerning the spatially-varying nature of warping problems \cite{bracewell1993affine}, frequency responses of an input and a deformed image are inconsistent for each location. Therefore, we formalize a Local Texture Estimator Warp (LTEW), a frequency response estimator for image warping, by considering both Fourier features of an input image and spatially-varying property of coordinate transformations. We show that our LTEW is a generalized form of the LTE, allowing a neural representation to be biased towards learning high-frequency components while manipulating images under arbitrary coordinate transformations. In addition, we present shape-dependent phase estimation to enrich the information in output images.

\subsection{Learning Fourier information for local neural representation}
In local neural representation \cite{Local_Implicit_Grid_CVPR20, chen2021learning}, a neural representation $g_\theta$ is parameterized by an MLP with trainable parameters $\theta$. A decoding function $g_\theta$ predicts RGB value for a query point $\mathbf y=f(\mathbf x)\in Y\subseteq\mathbb{R}^2$ as
\begin{gather}
    \mathbf {I^{WARP}}[\mathbf{y};\Theta]=\sum_{j\in\mathcal{J}}w_j g_\theta(\mathbf{z}_j, \mathbf{y}-f(\mathbf{x}_j))\label{eq:one}\\
    \mbox{where }\mathbf{z}=E_\varphi(\mathbf{I^{IN}})\label{eq:two},
\end{gather}
$\Theta=[\theta,\varphi]$, $\mathcal J$ is a set $\{j|j=[f^{-1}(\mathbf y)+[\frac{m}{w}, \frac{n}{h}], [m,n]\in[-1, 1]\}$ \cite{chen2021learning, NIPS2015_33ceb07b}, $w_j$ is a local ensemble coefficient \cite{Local_Implicit_Grid_CVPR20, chen2021learning, lee2021local}, $\mathbf z_j\in\mathbb{R}^C$ indicates a latent variable for a index $j$, and $\mathbf x_j\in X\subseteq\mathbb{R}^2$ is a coordinate of $\mathbf z_j$.

Recent works \cite{DBLP:conf/icml/RahamanBADLHBC19, tancik2020fourfeat} have shown that a standard MLP structure suffers from learning high-frequency content. Lee \textit{et al.} \cite{lee2021local} modified the local neural representation in Eq.~\eqref{eq:one} to overcome this spectral bias problem as
\begin{equation}
    \mathbf {I^{WARP}}[\mathbf{y};\Theta,\psi]=\sum_{j\in\mathcal{J}}w_j g_\theta(h_\psi(\mathbf{z}_j, \mathbf{y}-f(\mathbf{x}_j))\label{eq:three}
\end{equation}
where $h_\psi$ is a Local Texture Estimator (LTE). LTE ($h_\psi(\cdot)$) contains two estimators;(1) an amplitude estimator ($h_a(\cdot):\mathbb{R}^C\mapsto\mathbb{R}^{2D}$) (2) a frequency estimator ($h_f(\cdot):\mathbb{R}^C\mapsto\mathbb{R}^{2\times D}$). Specifically, an estimating function $h_\psi(\cdot,\cdot):(\mathbb R^C, \mathbb R^2)\mapsto\mathbb R^{2D}$ is defined as
\begin{gather}
h_\psi(\mathbf{z}_j,\boldsymbol{\delta}_{\mathbf y})=
\mathbf{A}_j\odot\begin{bmatrix}
\cos(\pi <\mathbf F_{j}, \boldsymbol{\delta}_{\mathbf y}>)\\
\sin(\pi <\mathbf F_{j}, \boldsymbol{\delta}_{\mathbf y}>)
\end{bmatrix},\label{eq:four}\\
\mbox{where }\mathbf{A}_j=h_a(\mathbf z_j),~
\mathbf{F}_j=h_f(\mathbf z_j),
\end{gather}
$\boldsymbol{\delta}_{\mathbf y}=\mathbf y - f(\mathbf x_j)$ is a local grid, $\mathbf A_j\in\mathbb R^{2D}$ is an amplitude vector, $\mathbf F_j\in\mathbb R^{2\times D}$ indicates a frequency matrix for an index $j$, $<\cdot,\cdot>$ is an inner product, and $\odot$ denotes element-wise multiplication. However, this formulation fails to represent warped images since $\mathbf F_j$ is a frequency response of an input image $\mathbf {I^{IN}}$, which is different from that of a warped image $\mathbf {I^{WARP}}$ \cite{bracewell1993affine}. In the following, we generalize LTE by considering a spatially-varying property of coordinate transformations.

\subsection{Learning Fourier information with  coordinate transformations}
We linearize the given coordinate transformation $f$ into affine transformations. A linear approximation of the local grid $\boldsymbol{\delta}_{\mathbf y}$ near a point $\mathbf x_j$ is computed as
\begin{equation}
\begin{aligned}
\boldsymbol{\delta}_{\mathbf y}
&=\mathbf y - f(\mathbf x_j)=f(\mathbf x) - f(\mathbf x_j)\\
&=\{f(\mathbf x_j) + \mathbf J_f(\mathbf x_j)(\mathbf x - \mathbf x_j) + \cancel{\mathcal{O}(\mathbf x^2)} \} - f(\mathbf x_j)\\
&\simeq \mathbf J_f(\mathbf x_j)(\mathbf x - \mathbf x_j) = \mathbf J_f(\mathbf x_j)\boldsymbol{\delta}_{\mathbf x}
\label{eq:six}
\end{aligned}
\end{equation}
where $\mathbf J_{f}(\mathbf x_j)\in\mathbb{R}^{2\times 2}$ is the Jacobian matrix of coordinate transformation $f$ at $\mathbf x_j$, $\mathcal{O}(\mathbf x^2)$ means terms of order $\mathbf x^2$ and higher, and $\boldsymbol{\delta}_{\mathbf x}=\mathbf x - \mathbf x_j$ is a local grid in input space $X$. By the affine theorem \cite{bracewell1993affine} and Eq.~\eqref{eq:six}, a frequency response $\mathbf F'_j$ of a warped image $\mathbf {I^{WARP}}$ near a point $f(\mathbf x_j)$ is approximated as follows:
\begin{equation}
\mathbf F'_j \simeq \mathbf J_f^{-T}(\mathbf x_j)\mathbf F_j.
\label{eq:seven}
\end{equation}

\noindent
From Eq.~\eqref{eq:six} and Eq.~\eqref{eq:seven}, we generalize the estimating function in Eq.~\eqref{eq:four} as:
\begin{align}
h_\psi(\mathbf{z}_j,\boldsymbol{\delta}_{\mathbf y}, f)
&=\mathbf{A}_j\odot\begin{bmatrix}
\cos(\pi <\mathbf F'_{j}, \boldsymbol{\delta}_{\mathbf y}>)\\
\sin(\pi <\mathbf F'_{j}, \boldsymbol{\delta}_{\mathbf y}>)
\end{bmatrix}\\
&\simeq\mathbf{A}_j\odot\begin{bmatrix}
\cos(\pi <\mathbf J_f^{-T}({\mathbf x_j})\mathbf F_{j}, \mathbf J_{f}({\mathbf x_j})\boldsymbol{\delta}_{\mathbf x}>)\\
\sin(\pi <\mathbf J_f^{-T}({\mathbf x_j})\mathbf F_{j}, \mathbf J_{f}({\mathbf x_j})\boldsymbol{\delta}_{\mathbf x}>)
\end{bmatrix}\\
&=\mathbf{A}_j\odot\begin{bmatrix}
\cos(\pi <\mathbf F_{j}, \boldsymbol{\delta}_{\mathbf x}>)\\
\sin(\pi <\mathbf F_{j}, \boldsymbol{\delta}_{\mathbf x}>)
\end{bmatrix}\\
&=h_\psi(\mathbf{z}_j,\boldsymbol{\delta}_{\mathbf x}).\label{eq:ten}
\end{align}

When comparing Eq.~\eqref{eq:four} and Eq.~\eqref{eq:ten}, we see that LTEW representation is capable of extracting Fourier information for warped images by utilizing the local grid in input coordinate space $\boldsymbol{\delta}_{\mathbf x}$ instead of $\boldsymbol{\delta}_{\mathbf y}$.

\begin{figure}[t]
\centering
\includegraphics[scale = 0.177]{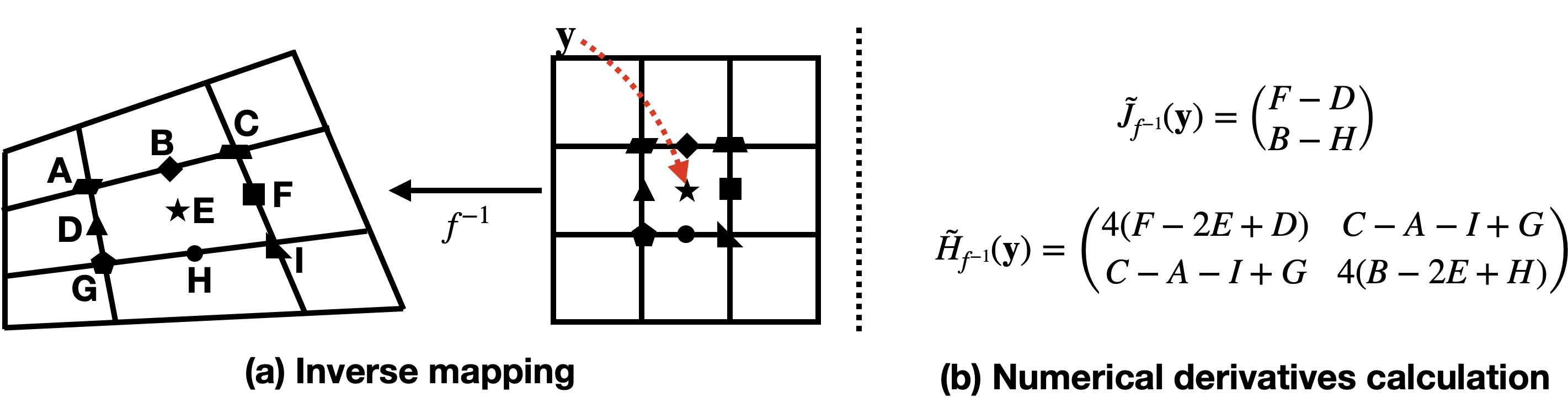}
\vspace{-6pt}
\caption{{Numerical Jacobian matrix and Hessian tensor for shape representation.}}
\label{fig:cell}
\end{figure}

\subsection{Shape-dependent phase estimation}
\label{sec:shpae}
For SISR tasks within symmetric scale factors, the pixel shape of upsampled images is square and spatially invariant. However, when it comes to image warping, pixels in resampled images are able to have arbitrary shapes and spatially-varying, as described in Fig.~\ref{fig:cell}.(a). To address this issue, we represent the pixel shape $\mathbf s(\mathbf y)\in\mathbb{R}^{12}$ ($\mathbb{R}^{4}$ is for pixel orientation, and $\mathbb{R}^{8}$ is for pixel curvature) with a gradient of coordinate transformation for a point $\mathbf y$ as
\begin{equation}
\boldsymbol{s}(\mathbf y)=[\tilde{\mathbf{J}}_{f^{-1}}({\mathbf y}), \tilde{\mathbf{H}}_{f^{-1}}({\mathbf y})]
\label{eq:cell}
\end{equation}
where $[\cdot, \cdot]$ refers to the concatenation after flattening, $\tilde{\mathbf{J}}_{f^{-1}}({\mathbf y})\in\mathbb{R}^{2\times 2}$ and $\tilde{\mathbf{H}}_{f^{-1}}({\mathbf y})\in\mathbb{R}^{2\times 2\times 2}$ denote the numerical Jacobian matrix, indicating an orientation of pixel, and the numerical Hessian tensor, specifying the degree of curvature, respectively. For shape representation, we apply an inverse coordinate transformation to a query point and its eight nearest points ($\mathbf y + [\frac{m}{W}, \frac{n}{H}]$ with $[m,n]\in[-1,0,1]$) and compute the difference to calculate numerical derivatives as described in Fig.~\ref{fig:cell}.(b). Let us assume that given coordinate transformation $f$ is in class $C^2$, which means $f_{x_1x_2}=f_{x_2x_1}$. Hence; we use only six elements in $\tilde{\mathbf H}_{f^{-1}}(\mathbf y)$ for shape representation.

Phase in Eq.~\eqref{eq:twelve} includes the information of edge locations or the shape of pixels \cite{lee2021local}. Therefore, we redefine the estimating function in Eq.~\eqref{eq:ten} as:
\begin{equation}
h_\psi(\mathbf{z}_j,\boldsymbol{\delta}_{\mathbf x}, \boldsymbol{s}(\mathbf y))
=\mathbf{A}_j\odot\begin{bmatrix}
\cos\{\pi (<\mathbf F_{j}, \boldsymbol{\delta}_{\mathbf x}>+h_p(\boldsymbol{s}(\mathbf y)))\}\\
\sin\{\pi (<\mathbf F_{j}, \boldsymbol{\delta}_{\mathbf x}>+h_p(\boldsymbol{s}(\mathbf y)))\}
\end{bmatrix}\label{eq:twelve}
\end{equation}
where $h_p(\cdot):\mathbb{R}^{10}\mapsto\mathbb{R}^{D}$ is a phase estimator.

Lastly, we add a bilinear interpolated image $\mathbf {I}_{\boldsymbol B}[\mathbf y]=\boldsymbol f_{\boldsymbol B}(\mathbf{I^{IN}})\simeq\mathbf {I^{IN}}(f^{-1}(\mathbf y))$ to stabilize network convergence and aid LTEW in learning high-frequency details \cite{Kim_2016_CVPR}. Thus, the local neural representation of a warped image $\mathbf {I^{WARP}}$ with the proposed LTEW is formulated as follows:
\begin{equation}
\mathbf {I^{WARP}}[\mathbf{y};\Theta,\psi]=\mathbf I_{\boldsymbol B}[\mathbf y]
+\sum_{j\in\mathcal{J}}w_j g_\theta(h_{\psi}(\mathbf{z}_j, f^{-1}(\mathbf{y})-\mathbf{x}_j, \boldsymbol{s}(\mathbf{y})))
\end{equation}

\begin{figure}[t]
\centering
\includegraphics[scale = 0.177]{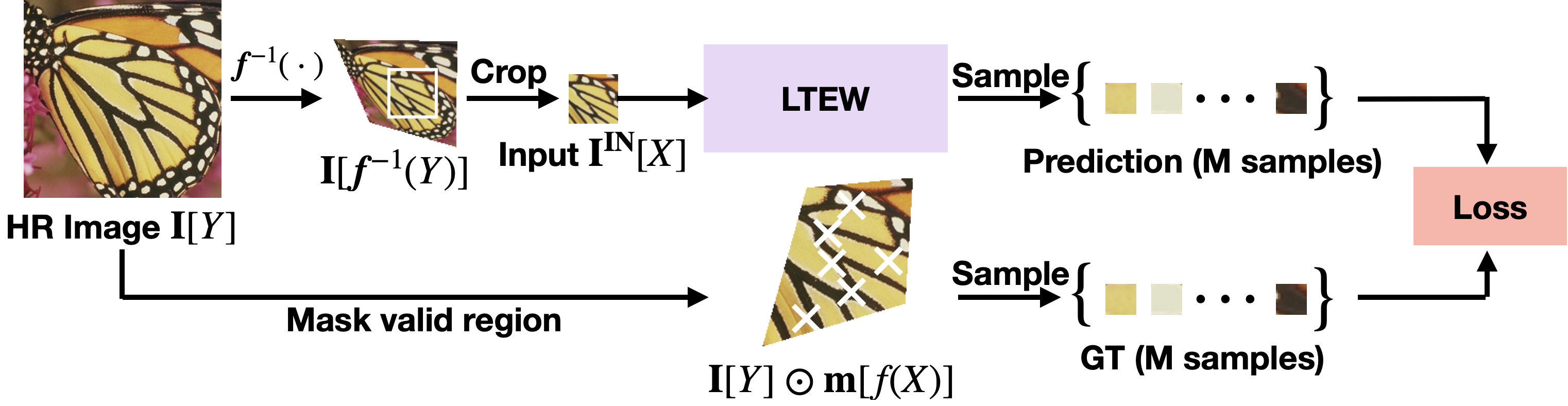}
\vspace{-6pt}
\caption{{Training strategy for Local Texture Estimator Warp (LTEW).}}
\label{fig:training}
\end{figure}

\section{Methods}
\subsection{Architecture details}
Our LTEW-based image warping network consists of an encoder ($E_\varphi$), the LTEW ($h_\psi$, a purple shaded region in Fig.~\ref{fig:concept}), and a decoder ($g_\theta$). An encoder ($E_\varphi$) is designed with a deep SR network, such as EDSR \cite{Lim_2017_CVPR_Workshops}, RCAN \cite{zhang2018rcan}, RRDB \cite{Wang_2018_ECCV_Workshops}, without upscaling modules. A decoder ($g_\theta$) is a 4-layer MLP with ReLUs, and its hidden dimension is 256. Our LTEW ($h_\psi$) takes a local grid ($\boldsymbol{\delta}_{\mathbf x}$), shape ($\mathbf s$), and feature map ($\mathbf z$) as input, and includes an amplitude estimator ($h_a$), a frequency estimator ($h_f$), and a phase estimator ($h_p$). An amplitude and a frequency estimator are implemented with $3\times 3$ convolution layer having 256 channels, and a phase estimator is a single linear layer with 128 channels. We assume that a warped image has the same texture near point $f(\mathbf x_j)$. Hence, we find estimated Fourier information ($\boldsymbol A_j, \boldsymbol F_j$) at $\mathbf x_j$ using the nearest-neighborhood interpolation. Then, estimated phase is added to an inner product between the local grid ($\boldsymbol \delta_{\mathbf x}$) and estimated frequencies, as in Eq.~\eqref{eq:twelve}. Before the decoder ($g_\theta$) resamples images, we multiply amplitude and sinusoidal activation output.

\subsection{Training strategy}
We have two batches with a size $\boldsymbol B$: (1) Image batch $\{I_{\boldsymbol 1} , I_{\boldsymbol 2}, \dots, I_{\boldsymbol B}\}$, $I_{\boldsymbol i}[Y]\in\mathbb{R}^{H\times W\times 3}$. (2) Coordinate transformation batch $\{f_{\boldsymbol 1}, f_{\boldsymbol 2}, \dots, f_{\boldsymbol B}\}$, where each $f_{\boldsymbol i}:X\mapsto Y$ is differentiable and invertible. For an input image preparation, we first apply an inverse coordinate transformation as:
\begin{equation}
\{I_{\boldsymbol 1}[f_{\boldsymbol 1}^{-1}(Y)] , I_{\boldsymbol 2}[f_{\boldsymbol 2}^{-1}(Y)], \dots, I_{\boldsymbol B}[f_{\boldsymbol B}^{-1}(Y)]\}
\end{equation}
where $I_{\boldsymbol i}[f_{\boldsymbol i}^{-1}(Y)]\in\mathbb{R}^{h_i\times w_i\times 3}$.
While avoiding void pixels, we crop input images $I_{\boldsymbol i}^{crop}[f_{\boldsymbol i}^{-1}(Y)]\in\mathbb{R}^{h\times w\times 3}$, where $h\leq\min(\{h_{\boldsymbol i}\}), w\leq\min(\{w_{\boldsymbol i}\})$. For a ground truth (GT) preparation, we randomly sample $\boldsymbol M$ query points among valid coordinates for an $\boldsymbol i$-th batch element as: $\{\mathbf y^{\boldsymbol i}_{\boldsymbol 1}, \mathbf y^{\boldsymbol i}_{\boldsymbol 2}, \dots, \mathbf y^{\boldsymbol i}_{\boldsymbol M}\}$, where $\mathbf y^{\boldsymbol i}_{\boldsymbol j}\in\mathbb{R}^2$. Then, we evaluate our LTEW for each query point with cropped input images to compute loss as in Fig.~\ref{fig:training} by comparing with following GT batch:
\begin{equation}
\begin{pmatrix}
    I_{\boldsymbol 1}[\mathbf y^{\boldsymbol 1}_{\boldsymbol 1}] & \dots  & I_{\boldsymbol B}[\mathbf y^{\boldsymbol B}_{\boldsymbol 1}] \\
    \vdots & \ddots & \\
    I_{\boldsymbol 1}[\mathbf y^{\boldsymbol 1}_{\boldsymbol M}] &        & I_{\boldsymbol B}[\mathbf y^{\boldsymbol B}_{\boldsymbol M}]
\end{pmatrix}
\end{equation}
where $I_{\boldsymbol i}[\mathbf y^{\boldsymbol i}_{\boldsymbol j}]$ is an RGB value.

For asymmetric-scale SR, each scale factor $s_x, s_y$ is randomly sampled from $\mathcal{U}(0.25,1)$. For homography transform, we randomly sample inverse coordinate transformation from the following distribution, dubbed \textit{in-scale}:
\begin{equation}
\begin{pmatrix}
    1   & h_x & 0 \\
    h_y & 1   & 0 \\
    0   & 0   & 1 \\
\end{pmatrix}
\begin{pmatrix}
    \cos{\theta}  & \sin{\theta} & 0 \\
    -\sin{\theta} & \cos{\theta} & 0 \\
    0   & 0   & 1 \\
\end{pmatrix}
\begin{pmatrix}
    s_x & 0   & 0 \\
    0   & s_y & 0 \\
    0   & 0   & 1 \\
\end{pmatrix}
\begin{pmatrix}
    1   & 0   & t_x \\
    0   & 1   & t_y \\
    p_x & p_y & 1 \\
\end{pmatrix}
\end{equation}
where $h_x, h_y\sim\mathcal{U}(-0.25,0.25)$ are for sheering, $\theta\sim\mathcal{N}(0,0.15^{{\circ}^2})$ is for rotation, $s_x, s_y\sim\mathcal{U}(0.35,0.5)$ are for scaling, $t_x\sim\mathcal{U}(-0.75W,0.125W)$, $t_y\sim\mathcal{U}(-0.75H,0.125H)$, $p_x\sim\mathcal{U}(-0.6W,0.6W)$, $p_y\sim\mathcal{U}(-0.6H,0.6H)$ are for projection. We evaluate our LTEW for unseen transformations to verify the generalization ability. For asymmetric-scale SR and homography transform, untrained coordinate transformations are sampled from $s_x, s_y\sim\mathcal{U}(0.125,0.25)$, dubbed \textit{out-of-scale}, other parameter distributions for $h_x, h_y, \theta, t_x, t_y, p_x, p_y$ remain the same.

\section{Experiments}
\subsection{Dataset and Training}
We use a DIV2K dataset \cite{8014884} of an NTIRE 2017 challenge \cite{8014883} for training. For optimization, we use an L1 loss \cite{Lim_2017_CVPR_Workshops} and an Adam \cite{DBLP:journals/corr/KingmaB14} with $\beta_1=0.9$, $\beta_2=0.999$. Networks are trained for 1000 epochs with batch size 16. The learning rate is initialized as 1e-4 and reduced by 0.5 at [200, 400, 600, 800]. Due to the page limit, evaluation details are provided in the supplementary.

\subsection{Evaluation}
\begin{figure*}[ht!]
\vspace{-12pt}
\footnotesize
\centering

\scriptsize{\raisebox{0.07in}{\rotatebox{90}{$\times1.6$/$\times3.8$}}}
\includegraphics[width=0.94in, height=0.705in]{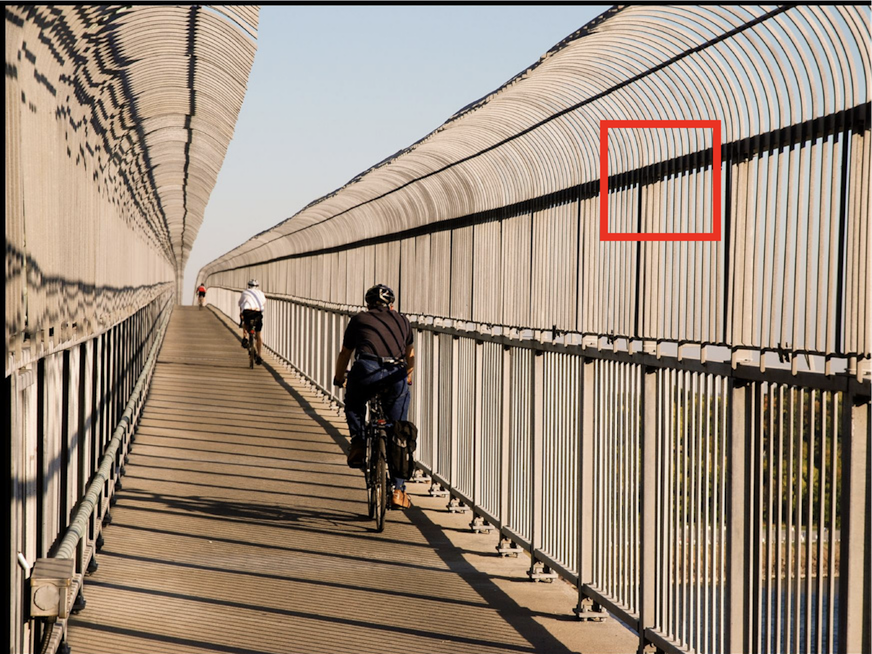}
\includegraphics[width=0.705in]{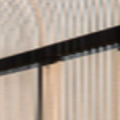}
\includegraphics[width=0.705in]{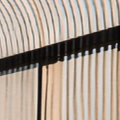}
\includegraphics[width=0.705in]{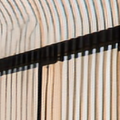}
\includegraphics[width=0.705in]{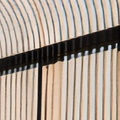}
\includegraphics[width=0.705in]{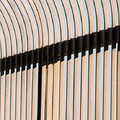}

\scriptsize{\raisebox{0.00in}{\rotatebox{90}{$\times3.55$/$\times1.55$}}}
\stackunder[2pt]{\includegraphics[width=0.94in, height=0.705in]{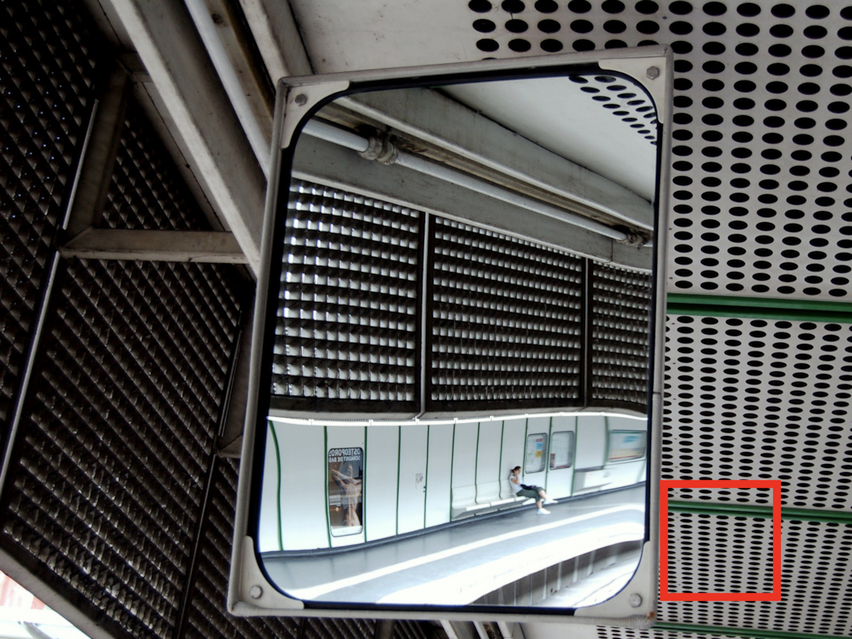}}{HR Image}
\stackunder[2pt]{\includegraphics[width=0.705in]{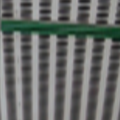}}{Bicubic}
\stackunder[2pt]{\includegraphics[width=0.705in]{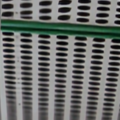}}{MetaSR \cite{hu2019meta}}
\stackunder[2pt]{\includegraphics[width=0.705in]{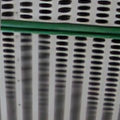}}{ArbSR \cite{Wang2020Learning}}
\stackunder[2pt]{\includegraphics[width=0.705in]{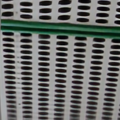}}{\textbf{LTEW}}
\stackunder[2pt]{\includegraphics[width=0.705in]{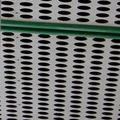}}{GT}



\vspace*{-6pt}
\caption{Qualitative comparison to other \underline{\textbf{asymmetric-scale SR}} within \underline{\textbf{in-scale}}. RCAN \cite{zhang2018rcan} is used as an encoder for all methods.}
\label{fig:Qual_asp}
\vspace{-12pt}
\end{figure*}

\begin{figure*}[ht!]
\vspace{-12pt}
\footnotesize
\centering

\scriptsize{\raisebox{0.15in}{\rotatebox{90}{$\times8$/$\times4$}}}
\includegraphics[width=0.94in, height=0.705in]{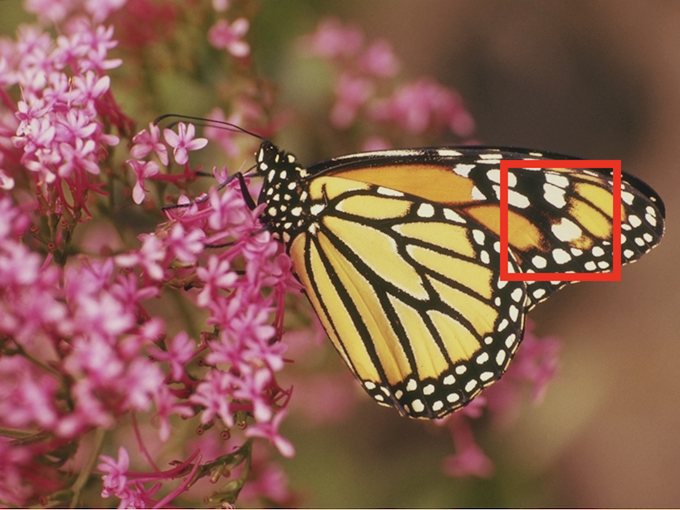}
\includegraphics[width=0.705in]{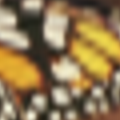}
\includegraphics[width=0.705in]{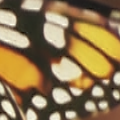}
\includegraphics[width=0.705in]{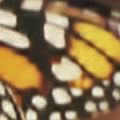}
\includegraphics[width=0.705in]{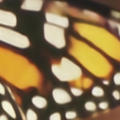}
\includegraphics[width=0.705in]{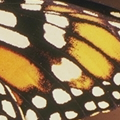}

\scriptsize{\raisebox{0.07in}{\rotatebox{90}{$\times3.2$/$\times7.6$}}}
\stackunder[2pt]{\includegraphics[width=0.94in, height=0.705in]{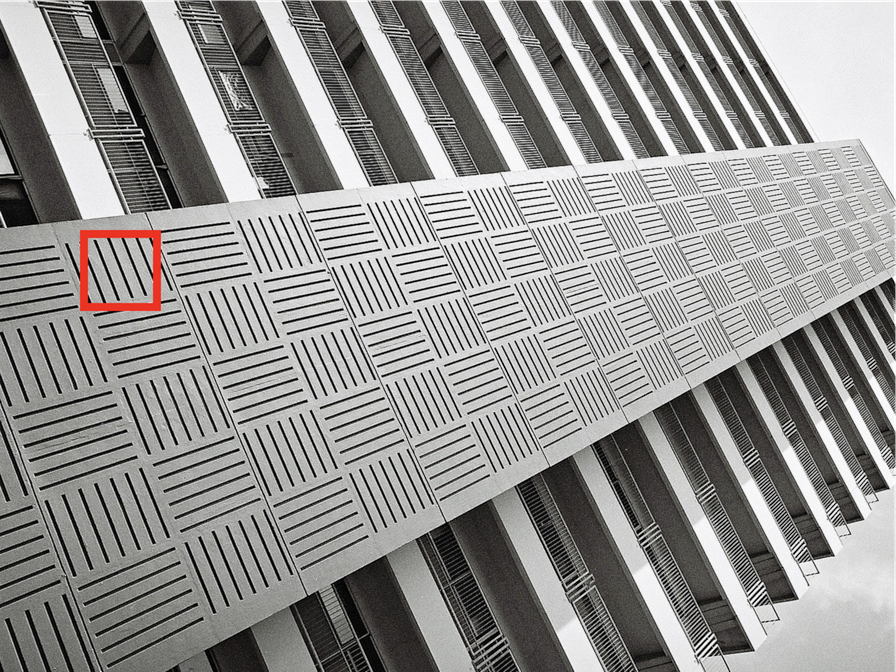}}{HR Image}
\stackunder[2pt]{\includegraphics[width=0.705in]{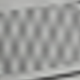}}{Bicubic}
\stackunder[2pt]{\includegraphics[width=0.705in]{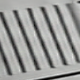}}{MetaSR \cite{hu2019meta}}
\stackunder[2pt]{\includegraphics[width=0.705in]{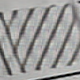}}{ArbSR \cite{Wang2020Learning}}
\stackunder[2pt]{\includegraphics[width=0.705in]{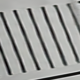}}{\textbf{LTEW}}
\stackunder[2pt]{\includegraphics[width=0.705in]{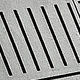}}{GT}

\vspace*{-6pt}
\caption{Qualitative comparison to other \underline{\textbf{asymmetric-scale SR}} within \underline{\textbf{out-of-scale}}. RCAN \cite{zhang2018rcan} is used as an encoder for all methods.}
\label{fig:Qual_asp_out_scale}
\end{figure*}

\textbf{Asymmetric-scale SR} We compare our LTEW for asymmetric-scale SR to RCAN \cite{zhang2018rcan}, MetaSR \cite{hu2019meta}, ArbSR \cite{Wang2020Learning} within \textit{in-scale} in Table~\ref{tab:Quan_Asymm}, Fig.~\ref{fig:Qual_asp} and \textit{out-of-scale} in Table~\ref{tab:Quan_Asymm_Out_scale}, Fig.~\ref{fig:Qual_asp_out_scale}. For RCAN \cite{zhang2018rcan}, we first upsample LR images by a factor of 4 and resample using bicubic interpolation. For MetaSR \cite{hu2019meta}, we first upsample input images by a factor of $\max(s_x, s_y)$ and downsample using a bicubic method as in \cite{Wang2020Learning}. Except for the case of Set14 $(\times3.5,\times2)$ and B100 $(\times1.5,\times3)$, LTEW outperforms existing methods within asymmetric scale factors in performance and visual quality for all scale factors and all datasets.

\begin{figure*}[ht!]
\vspace{-12pt}
\footnotesize
\centering

\includegraphics[width=0.96in, height=0.72in]{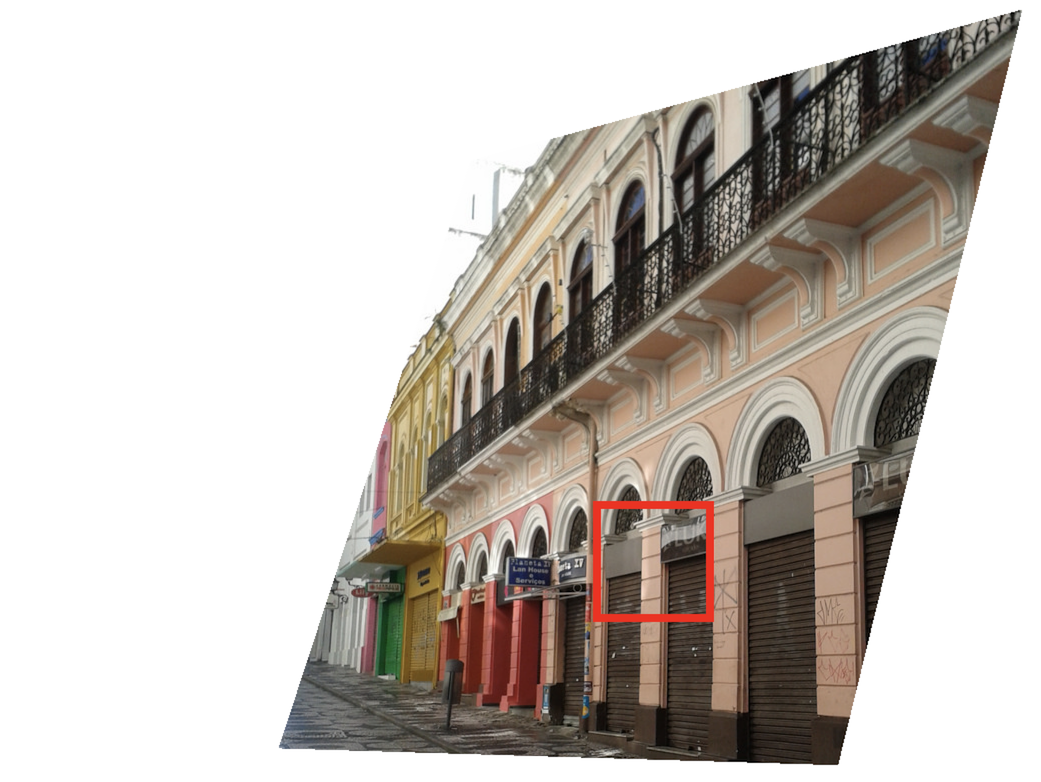}
\includegraphics[width=0.72in]{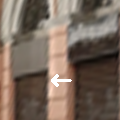}
\includegraphics[width=0.72in]{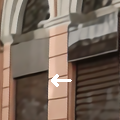}
\includegraphics[width=0.72in]{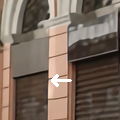}
\includegraphics[width=0.72in]{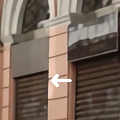}
\includegraphics[width=0.72in]{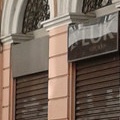}

\stackunder[2pt]{\includegraphics[width=0.96in, height=0.72in]{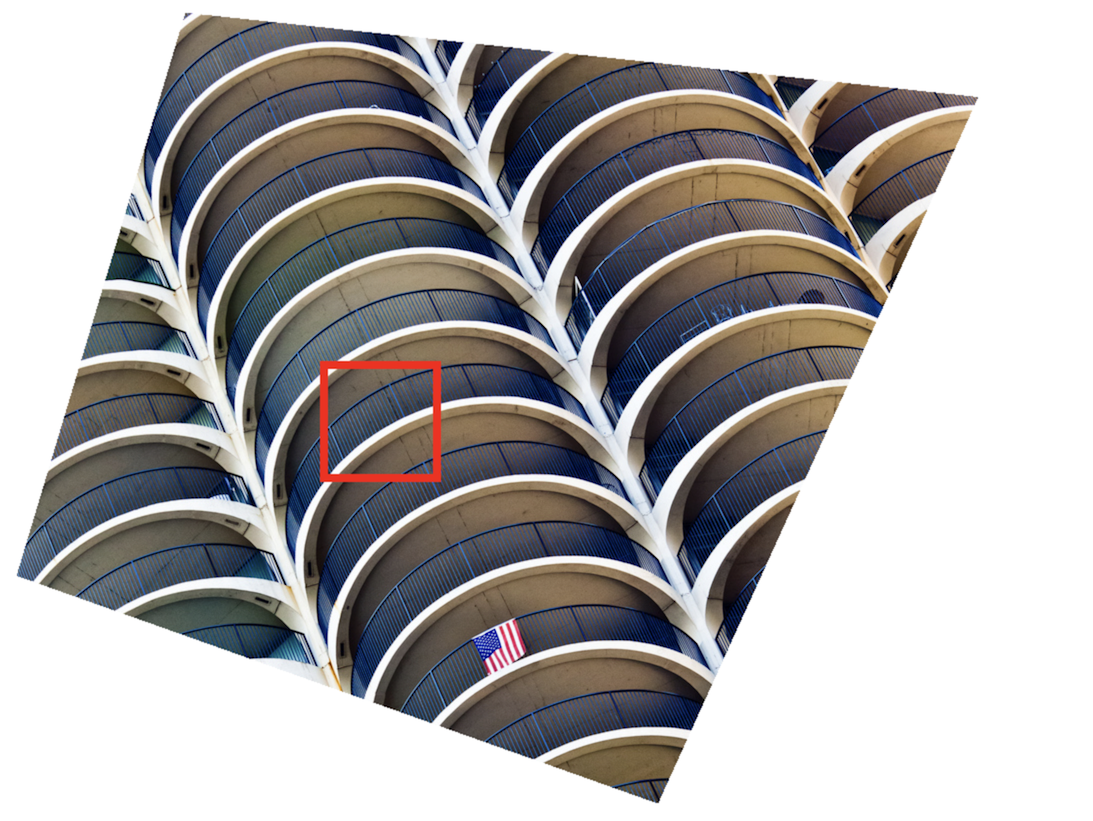}}{HR Image}
\stackunder[2pt]{\includegraphics[width=0.72in]{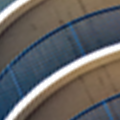}}{Bicubic}
\stackunder[2pt]{\includegraphics[width=0.72in]{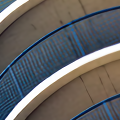}}{RRDB \cite{Wang_2018_ECCV_Workshops}}
\stackunder[2pt]{\includegraphics[width=0.72in]{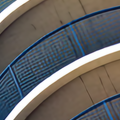}}{SRWarp \cite{SRWarp}}
\stackunder[2pt]{\includegraphics[width=0.72in]{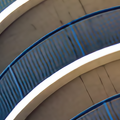}}{\textbf{LTEW}}
\stackunder[2pt]{\includegraphics[width=0.72in]{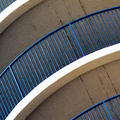}}{GT}

\vspace*{-6pt}
\caption{Qualitative comparison to other \underline{\textbf{homography transform}} within \underline{\textbf{in-scale}}. RRDB \cite{Wang_2018_ECCV_Workshops} is used as an encoder for all methods.}
\label{fig:Qual_warp}
\vspace{-12pt}
\end{figure*}

\begin{figure*}[ht!]
\vspace{-12pt}
\footnotesize
\centering

\includegraphics[width=0.96in, height=0.72in]{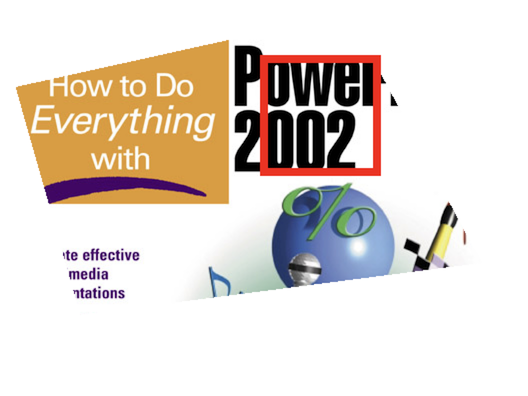}
\includegraphics[width=0.72in]{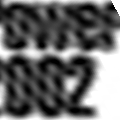}
\includegraphics[width=0.72in]{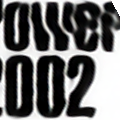}
\includegraphics[width=0.72in]{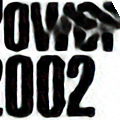}
\includegraphics[width=0.72in]{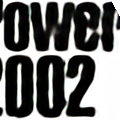}
\includegraphics[width=0.72in]{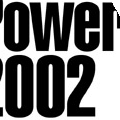}

\stackunder[2pt]{\includegraphics[width=0.96in, height=0.72in]{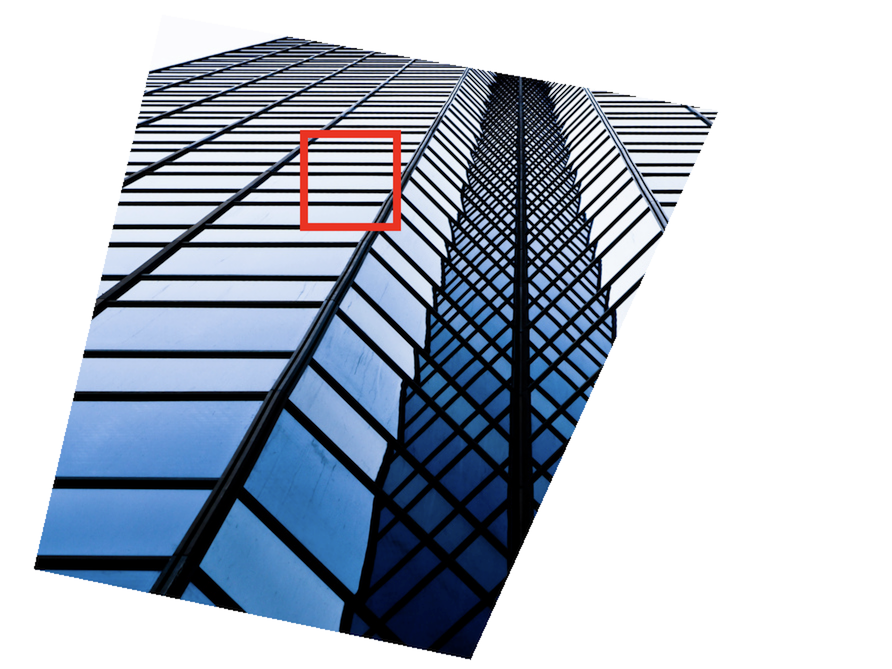}}{HR Image}
\stackunder[2pt]{\includegraphics[width=0.72in]{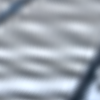}}{Bicubic}
\stackunder[2pt]{\includegraphics[width=0.72in]{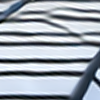}}{RRDB \cite{Wang_2018_ECCV_Workshops}}
\stackunder[2pt]{\includegraphics[width=0.72in]{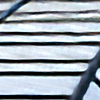}}{SRWarp \cite{SRWarp}}
\stackunder[2pt]{\includegraphics[width=0.72in]{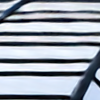}}{\textbf{LTEW}}
\stackunder[2pt]{\includegraphics[width=0.72in]{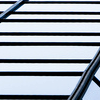}}{GT}

\vspace*{-6pt}
\caption{Qualitative comparison to other \underline{\textbf{homography transform}} within \underline{\textbf{out-of-scale}}. RRDB \cite{Wang_2018_ECCV_Workshops} is used as an encoder for all methods.}
\label{fig:Qual_warp_out_scale}
\vspace{-12pt}
\end{figure*}

\vspace{10pt}
\noindent
\textbf{Homography transform} We compare our LTEW for homography transform to RRDB \cite{Wang_2018_ECCV_Workshops} and SRWarp \cite{SRWarp} in Table~\ref{tab:Quan_Warp}, Fig.~\ref{fig:Qual_warp}, Fig.~\ref{fig:Qual_warp_out_scale}. For RRDB \cite{Wang_2018_ECCV_Workshops}, we super-sample input images by a factor of 4 and transform using bicubic resampling as in \cite{SRWarp}. We see that our LTEW surpasses existing homography transform methods in mPSNR and visual quality for both \textit{in-scale} and \textit{out-of-scale}

\newpage

\begin{table*}[ht!]
\centering
\setlength{\tabcolsep}{1.2pt}
\scriptsize{
\caption{Quantitative comparison with state-of-the-art methods for \underline{\textbf{asymmetric-scale SR}} within \underline{\textbf{in-scale}} on benchmark datasets (PSNR (dB)). \textcolor{red}{Red} and \textcolor{blue}{blue} colors indicate the best and the second-best performance, respectively.}
\vspace{-6pt}
\label{tab:Quan_Asymm}
\begin{tabular}{c
|>{\centering\arraybackslash}p{0.70cm}>{\centering\arraybackslash}p{0.70cm}>{\centering\arraybackslash}p{0.70cm}
|>{\centering\arraybackslash}p{0.70cm}>{\centering\arraybackslash}p{0.70cm}>{\centering\arraybackslash}p{0.70cm}
|>{\centering\arraybackslash}p{0.70cm}>{\centering\arraybackslash}p{0.70cm}>{\centering\arraybackslash}p{0.70cm}
|>{\centering\arraybackslash}p{0.70cm}>{\centering\arraybackslash}p{0.70cm}>{\centering\arraybackslash}p{0.70cm}}
\multirow{2}{*}{Method} & \multicolumn{3}{c|}{Set5} & \multicolumn{3}{c|}{Set14}
& \multicolumn{3}{c|}{B100} & \multicolumn{3}{c}{Urban100} \\
& $\frac{\times 1.5}{\times 4}$ & $\frac{\times 1.5}{\times 3.5}$ & $\frac{\times 1.6}{\times 3.05}$
& $\frac{\times 4}{\times 2}$   & $\frac{\times 3.5}{\times 2}$   & $\frac{\times 3.5}{\times 1.75}$
& $\frac{\times 4}{\times 1.4}$ & $\frac{\times 1.5}{\times 3}$   & $\frac{\times 3.5}{\times 1.45}$
& $\frac{\times 1.6}{\times 3}$ & $\frac{\times 1.6}{\times 3.8}$ & $\frac{\times 3.55}{\times 1.55}$ \\
\hline
Bicubic & 30.01 & 30.83 & 31.40
& 27.25 & 27.88 & 27.27
& 27.45 & 28.86 & 27.94
& 25.93 & 24.92 & 25.19\\
\hline
RCAN \cite{zhang2018rcan} & 34.14 & 35.05 & 35.67
& 30.35 & 31.02 & 31.21
& 29.35 & 31.30 & 29.98
& 30.72 & 28.81 & 29.34\\
MetaSR-RCAN \cite{hu2019meta} & 34.20 & 35.17 & 35.81
& 30.40 & 31.05 & 31.33
& 29.43 & 31.26 & 30.09
& 30.73 & 29.03 & 29.67\\
Arb-RCAN \cite{Wang2020Learning} & \textcolor{blue}{34.37} & \textcolor{blue}{35.40} & \textcolor{blue}{36.05}
& \textcolor{blue}{30.55} & \textcolor{red}{31.27} & \textcolor{blue}{31.54}
& \textcolor{blue}{29.54} & \textcolor{blue}{31.40} & \textcolor{blue}{30.22}
& \textcolor{blue}{31.13} & \textcolor{blue}{29.36} & \textcolor{blue}{30.04}\\
LTEW-RCAN (ours) & \textcolor{red}{34.45} & \textcolor{red}{35.46} & \textcolor{red}{36.12}
& \textcolor{red}{30.57} & \textcolor{blue}{31.21} & \textcolor{red}{31.55}
& \textcolor{red}{29.62} & \textcolor{blue}{31.40} & \textcolor{red}{30.24}
& \textcolor{red}{31.25} & \textcolor{red}{29.57} & \textcolor{red}{30.21}\\
\end{tabular}
}
\vspace{-12pt}
\end{table*}

\begin{table*}[ht!]
\vspace{-12pt}
\centering
\setlength{\tabcolsep}{1.2pt}
\scriptsize{
\caption{Quantitative comparison with state-of-the-art methods for \underline{\textbf{asymmetric-scale SR}} within \underline{\textbf{out-of-scale}} on benchmark datasets (PSNR (dB)). \textcolor{red}{Red} and \textcolor{blue}{blue} colors indicate the best and the second-best performance, respectively.}
\vspace{-6pt}
\label{tab:Quan_Asymm_Out_scale}
\begin{tabular}{c
|>{\centering\arraybackslash}p{0.70cm}>{\centering\arraybackslash}p{0.70cm}>{\centering\arraybackslash}p{0.70cm}
|>{\centering\arraybackslash}p{0.70cm}>{\centering\arraybackslash}p{0.70cm}>{\centering\arraybackslash}p{0.70cm}
|>{\centering\arraybackslash}p{0.70cm}>{\centering\arraybackslash}p{0.70cm}>{\centering\arraybackslash}p{0.70cm}
|>{\centering\arraybackslash}p{0.70cm}>{\centering\arraybackslash}p{0.70cm}>{\centering\arraybackslash}p{0.70cm}}
\multirow{2}{*}{Method} & \multicolumn{3}{c|}{Set5} & \multicolumn{3}{c|}{Set14}
& \multicolumn{3}{c|}{B100} & \multicolumn{3}{c}{Urban100} \\
& $\frac{\times 3}{\times 8}$ & $\frac{\times 3}{\times 7}$ & $\frac{\times 3.2}{\times 6.1}$
& $\frac{\times 8}{\times 4}$   & $\frac{\times 7}{\times 4}$   & $\frac{\times 7}{\times 3.5}$
& $\frac{\times 8}{\times 2.8}$ & $\frac{\times 3}{\times 6}$   & $\frac{\times 7}{\times 2.9}$
& $\frac{\times 3.2}{\times 6}$ & $\frac{\times 3.2}{\times 7.6}$ & $\frac{\times 7.1}{\times 3.1}$ \\
\hline
Bicubic & 25.69 & 26.35 & 26.84
& 24.27 & 24.62 & 24.79
& 24.67 & 25.58 & 24.98
& 22.55 & 21.92 & 22.15\\
\hline
RCAN \cite{zhang2018rcan} & \textcolor{blue}{29.00} & \textcolor{blue}{30.01} & \textcolor{blue}{30.46}
& \textcolor{blue}{26.48} & \textcolor{blue}{26.94} & \textcolor{blue}{27.11}
& \textcolor{black}{26.06} & \textcolor{blue}{27.19} & \textcolor{blue}{26.47}
& \textcolor{blue}{25.52} & \textcolor{blue}{24.50} & \textcolor{blue}{24.84}\\
MetaSR-RCAN \cite{hu2019meta} & \textcolor{black}{28.75} & \textcolor{black}{29.74} & \textcolor{black}{30.38}
& \textcolor{black}{26.32} & \textcolor{black}{26.85} & \textcolor{black}{27.03}
& \textcolor{blue}{26.07} & \textcolor{black}{27.15} & \textcolor{black}{26.45}
& \textcolor{black}{25.50} & \textcolor{black}{24.47} & \textcolor{blue}{24.84} \\
Arb-RCAN \cite{Wang2020Learning} & \textcolor{black}{28.37} & \textcolor{black}{29.35} & \textcolor{black}{30.08}
& \textcolor{black}{26.06} & \textcolor{black}{26.63} & \textcolor{black}{26.84}
& \textcolor{black}{25.91} & \textcolor{black}{27.14} & \textcolor{black}{26.40}
& \textcolor{black}{25.36} & \textcolor{black}{24.12} & \textcolor{black}{24.61}\\
LTEW-RCAN (ours) & \textcolor{red}{29.26} & \textcolor{red}{30.16} & \textcolor{red}{30.64}
& \textcolor{red}{26.60} & \textcolor{red}{27.06} & \textcolor{red}{27.25}
& \textcolor{red}{26.25} & \textcolor{red}{27.28} & \textcolor{red}{26.62}
& \textcolor{red}{25.85} & \textcolor{red}{24.79} & \textcolor{red}{25.18}\\
\end{tabular}
}
\vspace{-12pt}
\end{table*}

\newpage

\begin{table*}[ht]
\centering
\setlength{\tabcolsep}{1.2pt}
\scriptsize{
\caption{Quantitative comparison with state-of-the-art methods for \underline{\textbf{homography transform}} within \underline{\textbf{in-scale}} (\textit{isc}) and \underline{\textbf{out-of-scale}} (\textit{osc}) on DIV2KW and benchmarkW datasets (mPSNR (dB)). \textcolor{red}{Red} and \textcolor{blue}{blue} colors indicate the best and the second-best performance, respectively.}
\label{tab:Quan_Warp}
\vspace{-6pt}
\begin{tabular}{c
|>{\centering\arraybackslash}p{0.70cm}>{\centering\arraybackslash}p{0.70cm}
|>{\centering\arraybackslash}p{0.70cm}>{\centering\arraybackslash}p{0.70cm}
|>{\centering\arraybackslash}p{0.70cm}>{\centering\arraybackslash}p{0.70cm}
|>{\centering\arraybackslash}p{0.70cm}>{\centering\arraybackslash}p{0.70cm}
|>{\centering\arraybackslash}p{0.70cm}>{\centering\arraybackslash}p{0.70cm}}
\multirow{2}{*}{Method} & \multicolumn{2}{c|}{DIV2KW} & \multicolumn{2}{c|}{Set5W} & \multicolumn{2}{c|}{Set14W}
& \multicolumn{2}{c|}{B100W} & \multicolumn{2}{c}{Urban100W}\\
& \textit{isc} & \textit{osc} & \textit{isc} & \textit{osc} & \textit{isc} & \textit{osc} & \textit{isc} & \textit{osc} & \textit{isc} & \textit{osc}\\
\hline
Bicubic & 27.85 & \textcolor{black}{25.03} & 35.00 & \textcolor{black}{28.75} & 28.79 & \textcolor{black}{24.57} & 28.67 & \textcolor{black}{25.02} & 24.84 & \textcolor{black}{21.89} \\
\hline
RRDB \cite{Wang_2018_ECCV_Workshops} & 30.76 & \textcolor{blue}{26.84} & 37.40 & \textcolor{blue}{30.34} & 31.56 & \textcolor{blue}{25.95} & 30.29 & \textcolor{blue}{26.32} & 28.83 & \textcolor{black}{23.94}\\
SRWarp-RRDB \cite{SRWarp} & \textcolor{blue}{31.04} & \textcolor{black}{26.75} & \textcolor{blue}{37.93} & \textcolor{black}{29.90} & \textcolor{blue}{32.11} & \textcolor{black}{25.35} & \textcolor{blue}{30.48} & \textcolor{black}{26.10} & \textcolor{blue}{29.45} & \textcolor{blue}{24.04} \\
LTEW-RRDB (ours) & \textcolor{red}{31.10} & \textcolor{red}{26.92} & \textcolor{red}{38.20} & \textcolor{red}{31.07} & \textcolor{red}{32.15} & \textcolor{red}{26.02} & \textcolor{red}{30.56} & \textcolor{red}{26.41} & \textcolor{red}{29.50} & \textcolor{red}{24.25}
\end{tabular}
}
\vspace{-12pt}
\end{table*}

\begin{figure*}[t]
\footnotesize
\centering

\includegraphics[width=1.52in, height=0.76in]{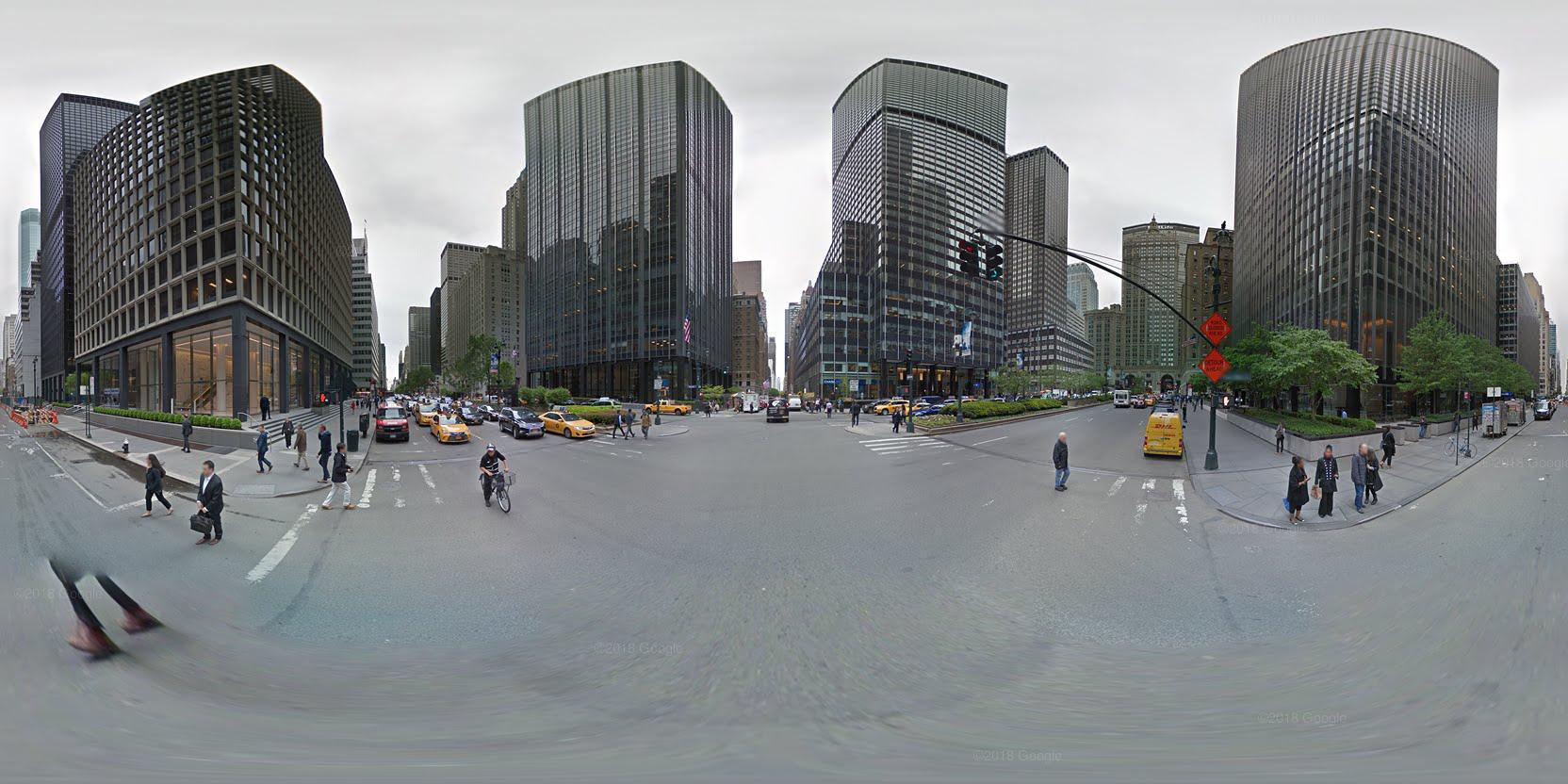}
\includegraphics[width=0.76in]{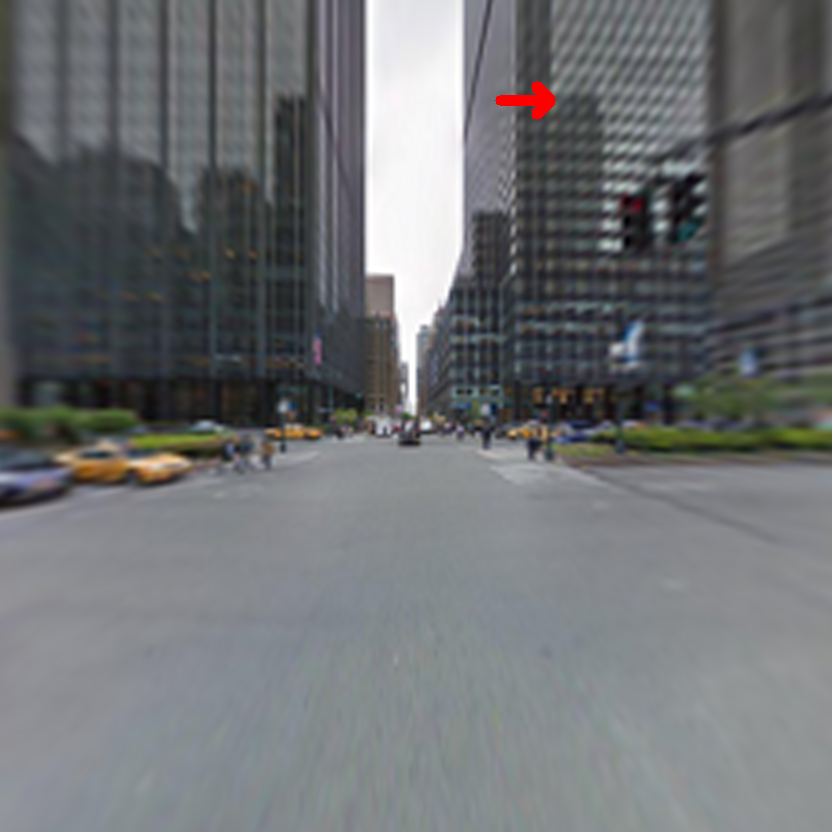}
\includegraphics[width=0.76in]{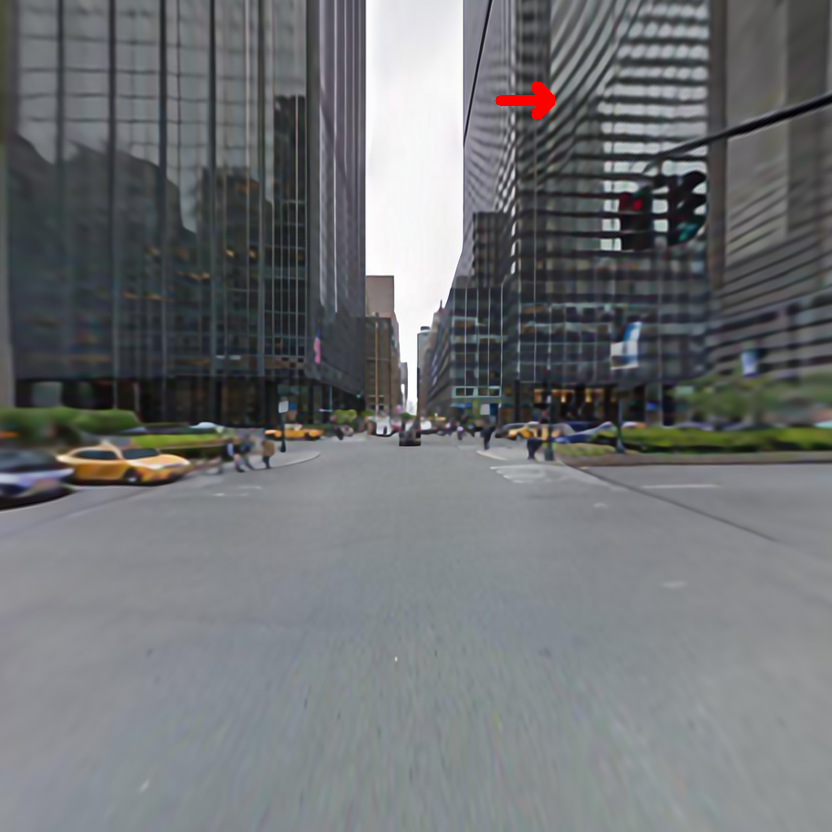}
\includegraphics[width=0.76in]{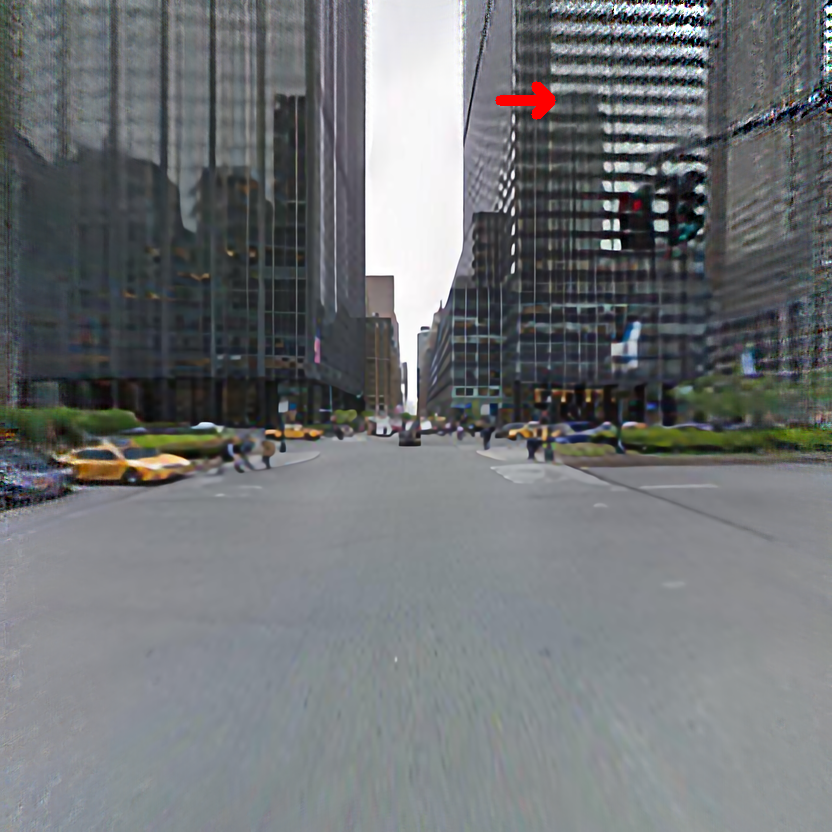}
\includegraphics[width=0.76in]{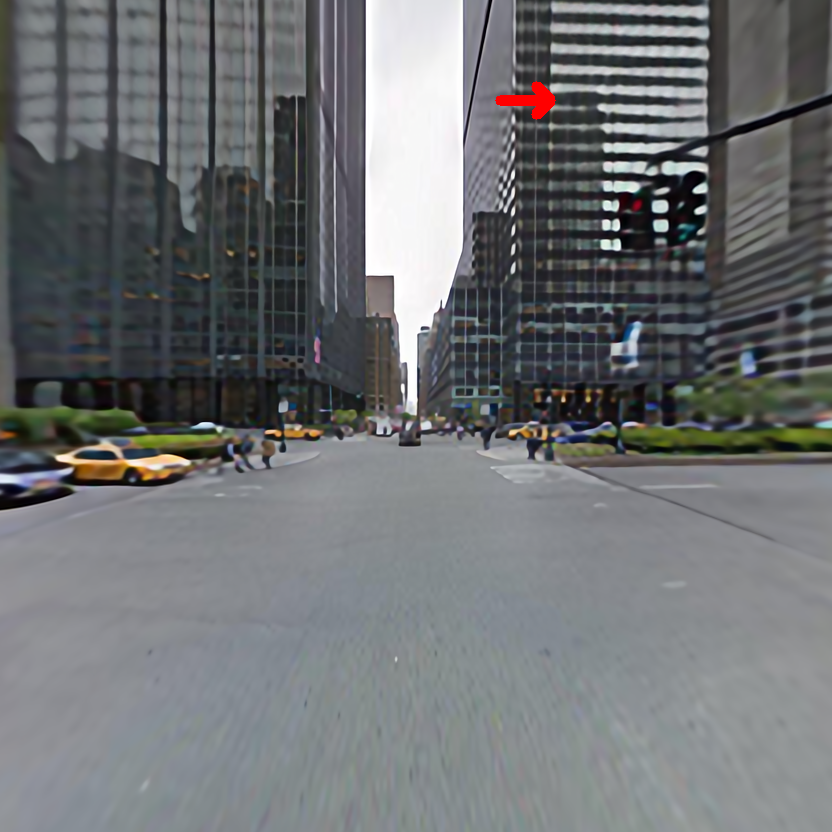}


\stackunder[2pt]{\includegraphics[width=1.52in, height=0.76in]{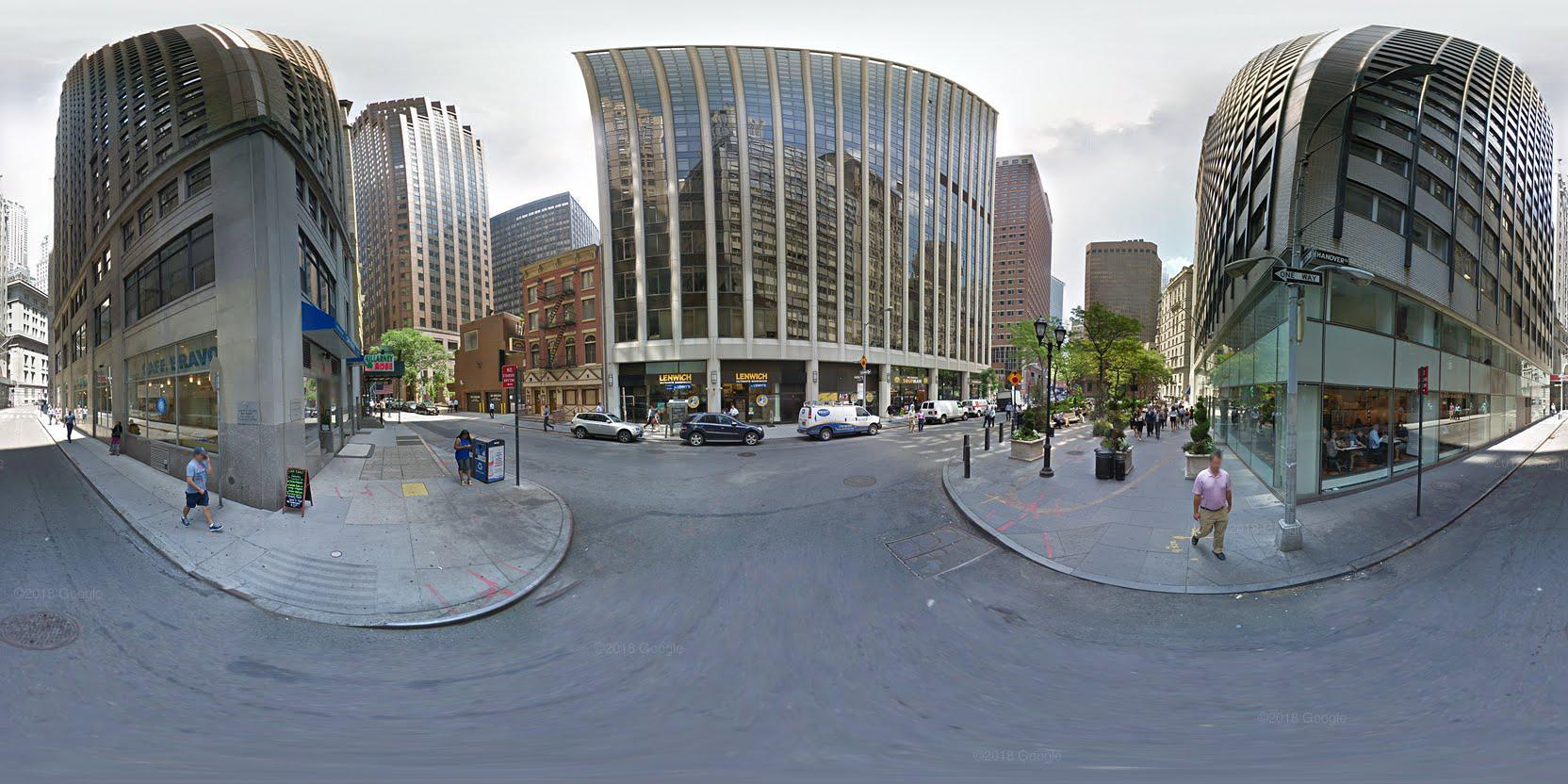}}{ERP Image}
\stackunder[2pt]{\includegraphics[width=0.76in]{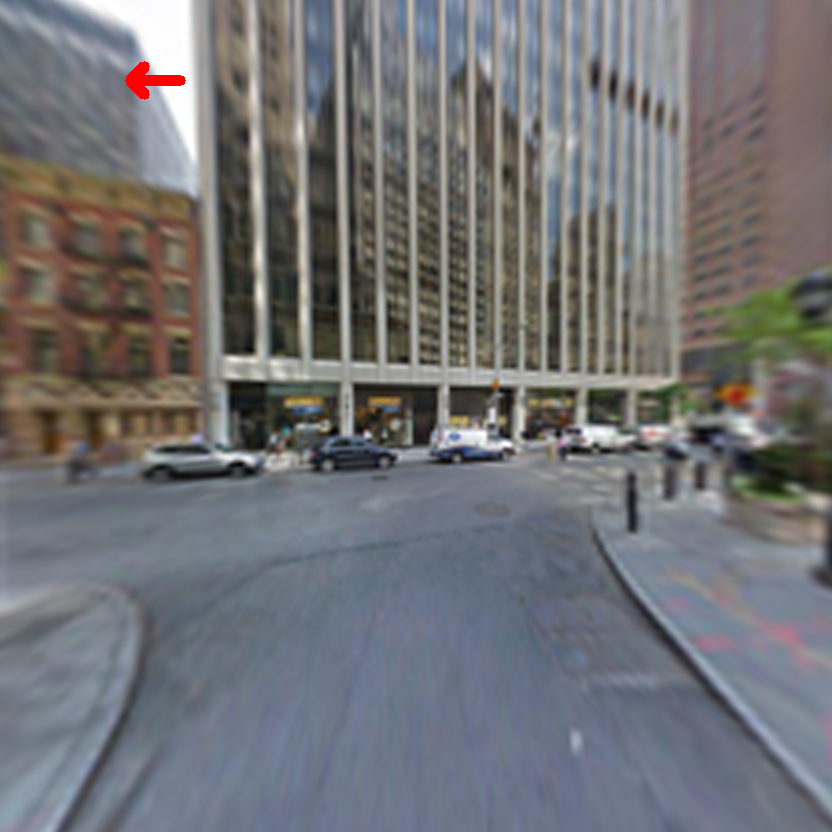}}{Bicubic}
\stackunder[2pt]{\includegraphics[width=0.76in]{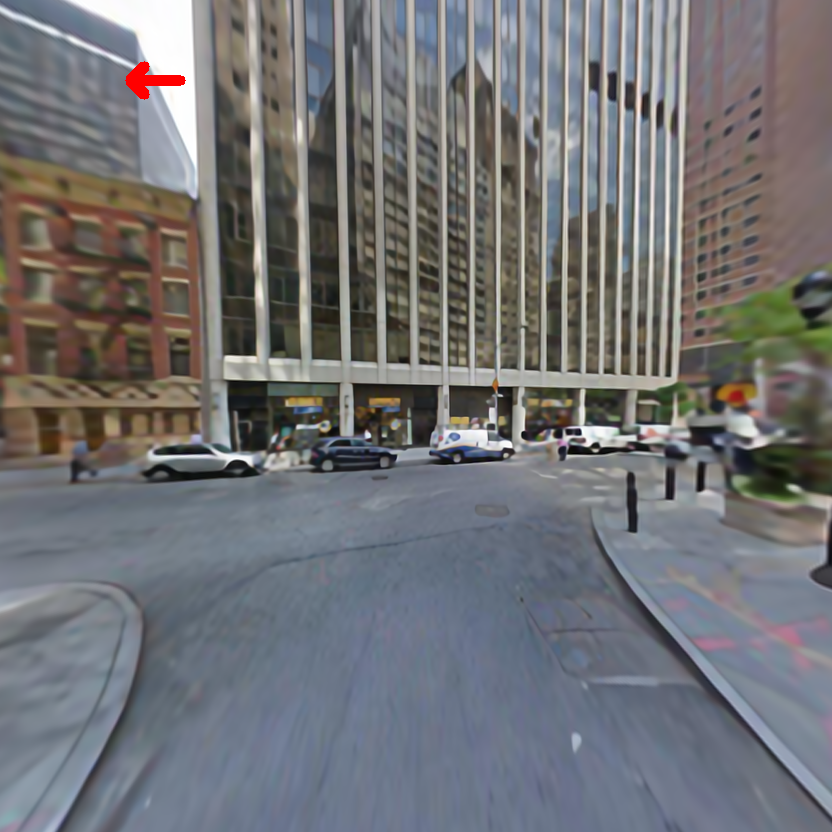}}{RRDB \cite{Wang_2018_ECCV_Workshops}}
\stackunder[2pt]{\includegraphics[width=0.76in]{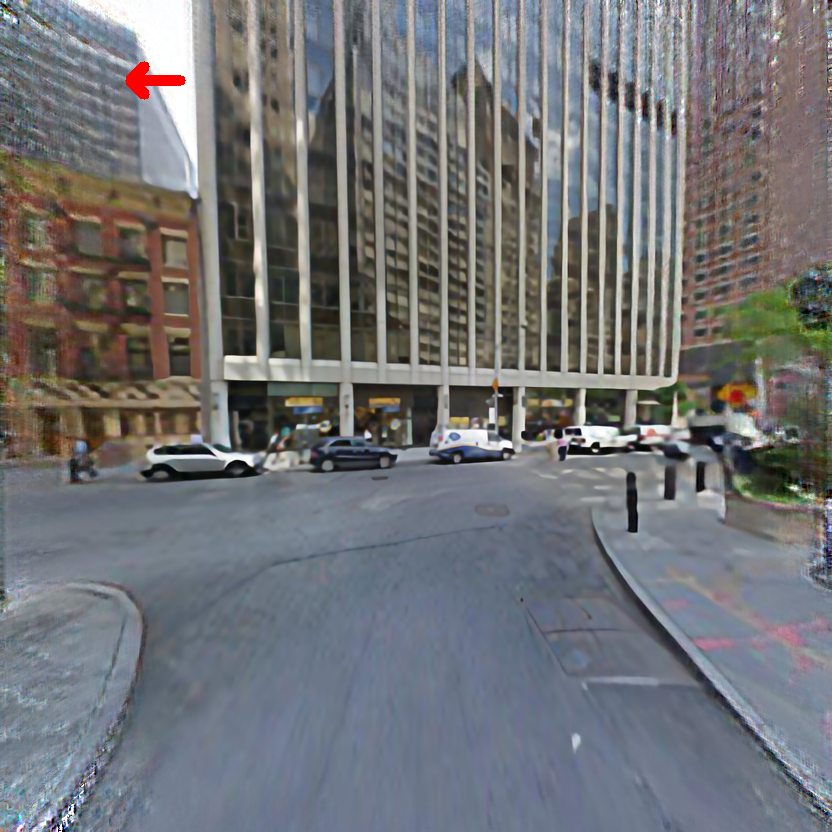}}{SRWarp \cite{SRWarp}}
\stackunder[2pt]{\includegraphics[width=0.76in]{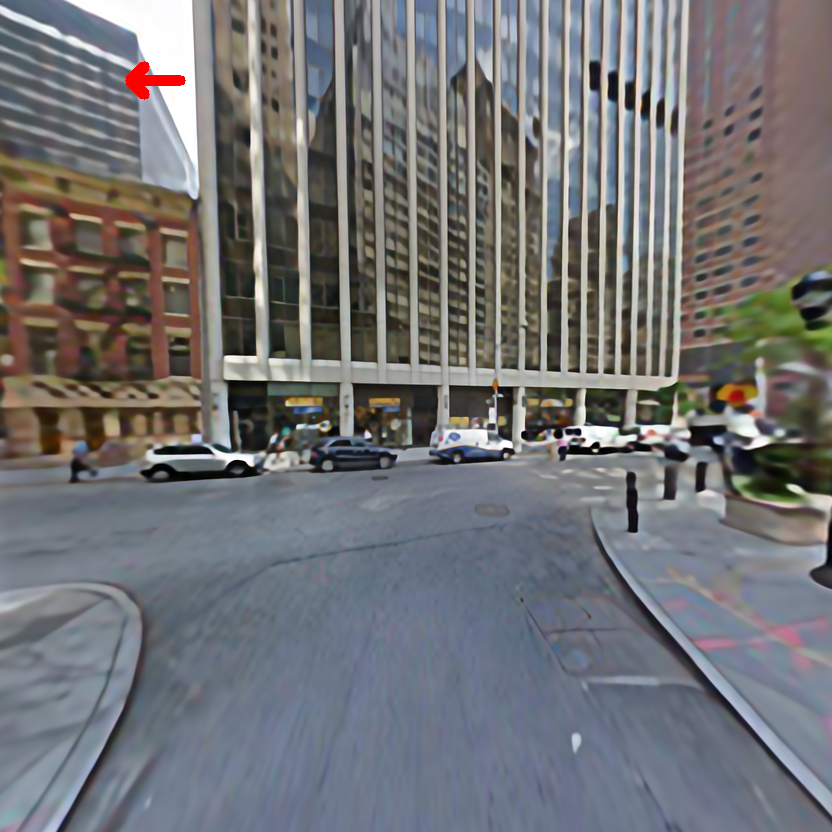}}{\textbf{LTEW}}

\vspace*{-6pt}
\caption{Qualitative comparison to other \underline{\textbf{warping}} for \underline{\textbf{ERP perspective projection}}. RRDB \cite{Wang_2018_ECCV_Workshops} is used as an encoder for all methods.}
\label{fig:Qual_warp_erp_correction}
\end{figure*}

\begin{figure*}[ht!]
\footnotesize
\centering

\stackunder[2pt]{\includegraphics[height=0.88in]{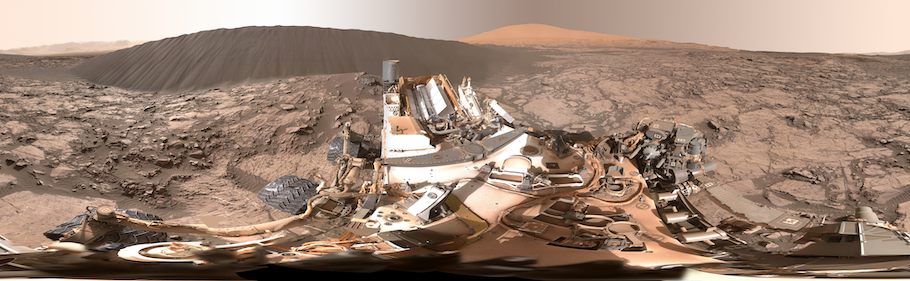}}{ERP Image (8190px $\times$ 2529px)}
\stackunder[2pt]{\includegraphics[height=0.88in]{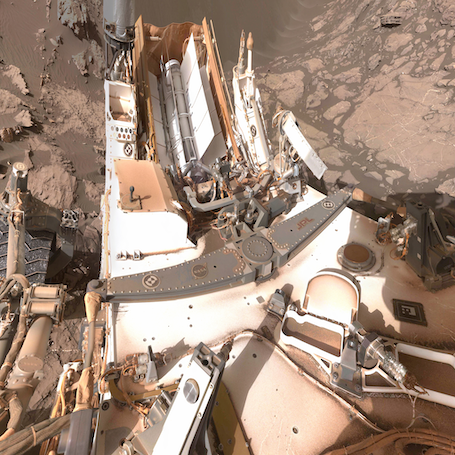}}{LTEW}
\stackunder[2pt]{\includegraphics[height=0.88in]{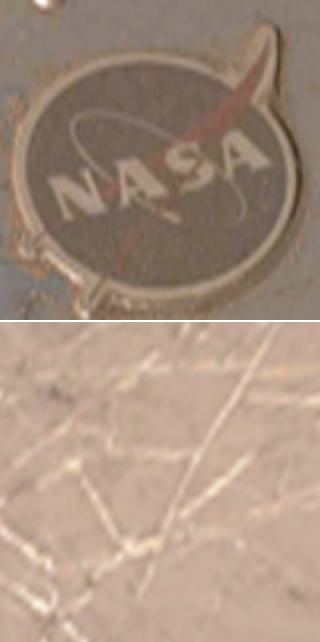}}{Bicubic}
\stackunder[2pt]{\includegraphics[height=0.88in]{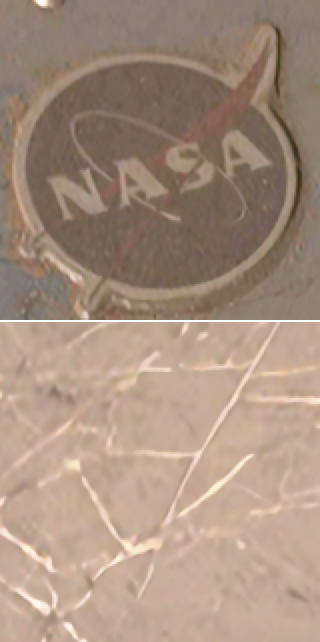}}{\textbf{LTEW}}
\vspace*{-6pt}
\caption{Visual demonstration of LTEW for \underline{\textbf{HR ERP perspective projection}}. RRDB \cite{Wang_2018_ECCV_Workshops} and SRWarp \cite{SRWarp} are not able to evaluate given ERP size (8190px$\times$2529px) under CPU-computing with 512GB RAM.}
\label{fig:Qual_mars_erp_correction}
\end{figure*}


\noindent
\textbf{ERP perspective projection} We train our LTEW to perform homography transform and apply an unseen transformation: ERP perspective projection, to validate the generalization ability. Since we are not able to obtain high-quality GT for ERP perspective projection (due to JPEG compression artifact), we visually compare our method to RRDB \cite{Wang_2018_ECCV_Workshops} and SRWarp \cite{SRWarp} in Fig.~\ref{fig:Qual_warp_erp_correction}. For RRDB \cite{Wang_2018_ECCV_Workshops}, we upsample input ERP images by a factor of 4 and interpolate them in a bicubic manner. Resolution of input ERP images is $1664\times 832$. As pointed out in \cite{Deng_2021_CVPR}, considering HMD's limited hardware resources, storing and transmitting ERP images in full resolution is impractical. Therefore, we downsample HR ERP images by a factor of 4 and then project images to a size of $832\times 832$ with a field of view (FOV) $120^{\circ}$. From Fig.~\ref{fig:Qual_warp_erp_correction}, we observe that RRDB \cite{Wang_2018_ECCV_Workshops} is limited in capturing high-frequency details, and SRWarp \cite{SRWarp} shows artifacts near boundaries. In contrast, our proposed LTEW captures fine details without any artifacts near boundaries.

In Fig.~\ref{fig:Qual_mars_erp_correction}, we project an HR ERP image with size $8190\times 2529$ to $4095\times 4095$ with a FOV $90^{\circ}$ on the Intel Xeon Gold 6226R@2.90GHz and RAM 512GB. Note that RRDB \cite{Wang_2018_ECCV_Workshops} and SRWarp \cite{SRWarp} are not able to evaluate given HR ERP size even under CPU-computing. Both RRDB and SRWarp need to upsample an ERP image by a factor of 4, consuming a massive amount of memory. In contrast, our LTEW is memory-efficient since we are able to query evaluation points sequentially. We notice that our LTEW is capable of restoring sharper and clearer edges compared to bicubic interpolation.

\subsection{Ablation study}
\label{sec:abl}

\begin{table*}[ht!]
\centering
\setlength{\tabcolsep}{1.2pt}
\scriptsize{
\caption{Quantitative ablation study for \underline{\textbf{homography transform}} within \underline{\textbf{in-scale}} (\textit{isc}) and \underline{\textbf{out-of-scale}} (\textit{osc}) on DIV2KW dataset (mPSNR (dB)). EDSR-baseline \cite{Lim_2017_CVPR_Workshops} is used as an encoder.}
\label{tab:Quan_Abl}
\vspace{-6pt}
\begin{tabular}{c
|>{\centering\arraybackslash}p{0.68cm}>{\centering\arraybackslash}p{0.68cm}>{\centering\arraybackslash}p{0.68cm}
|>{\centering\arraybackslash}p{0.80cm}
|>{\centering\arraybackslash}p{0.77cm}>{\centering\arraybackslash}p{0.77cm}
|>{\centering\arraybackslash}p{1.65cm}>{\centering\arraybackslash}p{1.65cm}}
Method & Amp. & Long. & $N_{F_j}$ & Act. & Jacob. & Hess. & \textit{isc} & \textit{osc}\\
\hline
\multicolumn{1}{l|}{LIIF ($-c$) \cite{chen2021learning}} & \multicolumn{3}{c|}{Concat} & ReLU & \xmark & \xmark & \textcolor{black}{30.65}($-$0.28) & \textcolor{black}{26.73}($+$0.00)\\
\multicolumn{1}{l|}{LIIF \cite{chen2021learning} $+$ Eq.~\eqref{eq:cell}} & \multicolumn{3}{c|}{Concat} & ReLU & \checkmark & \checkmark & \textcolor{black}{30.74}($-$0.09) & \textcolor{black}{26.66}($-$0.07)\\
\multicolumn{1}{l|}{LIIF \cite{chen2021learning} $+$ Eq.~\eqref{eq:cell} $+$ \cite{sitzmann2019siren}} & \multicolumn{3}{c|}{Concat} & $\sin$ & \checkmark & \checkmark & \textcolor{black}{30.49}($-$0.34) & \textcolor{black}{26.52}($-$0.21)\\
\multicolumn{1}{l|}{ITSRN ($-token$) \cite{itsrn}} & \multicolumn{3}{c|}{Transformer} & ReLU & \xmark & \xmark & \textcolor{black}{30.64}($-$0.19) & \textcolor{black}{26.69}($-$0.04)\\
\multicolumn{1}{l|}{ITSRN \cite{itsrn} $+$ Eq.~\eqref{eq:cell}} & \multicolumn{3}{c|}{Transformer} & ReLU & \checkmark & \checkmark & \textcolor{black}{30.74}($-$0.09) & \textcolor{black}{26.52}($-$0.21)\\
\hline
\multicolumn{1}{l|}{LTEW ($-A$)} & \xmark & \checkmark & 256 & ReLU & \checkmark & \checkmark & \textcolor{black}{30.77}($-$0.06) & \textcolor{black}{26.66}($-$0.07)\\
\multicolumn{1}{l|}{LTEW ($-L$)} & \checkmark & \xmark & 256 & ReLU & \checkmark & \checkmark & \textcolor{black}{30.79}($-$0.04) & \textcolor{black}{26.67}($-$0.06)\\
\multicolumn{1}{l|}{LTEW ($-F$)} & \checkmark & \checkmark & 128 & ReLU & \checkmark & \checkmark & \textcolor{black}{30.80}($-$0.03) & \textcolor{black}{26.70}($-$0.03)\\
\hline
\multicolumn{1}{l|}{LTEW ($-P_{J}$, $-P_{H}$, $+c$\cite{lee2021local})} & \checkmark & \checkmark & 256 & ReLU & \multicolumn{2}{c|}{$(2r_\mathbf{x}, 2/r_\mathbf{y})$}  & \textcolor{black}{25.23}($-$5.60) & \textcolor{black}{25.91}($-$0.82)\\
\multicolumn{1}{l|}{LTEW ($-P_{J}$, $-P_{H}$)} & \checkmark & \checkmark & 256 & ReLU & \xmark & \xmark & \textcolor{black}{30.69}($-$0.14) & \textbf{\textcolor{black}{26.76}}($+$0.03)\\
\multicolumn{1}{l|}{LTEW ($-P_{J}$)} & \checkmark & \checkmark & 256 & ReLU & \xmark & \checkmark & \textcolor{black}{30.70}($-$0.13) & \textcolor{black}{26.72}($-$0.01)\\
\multicolumn{1}{l|}{LTEW ($-P_{H}$)} & \checkmark & \checkmark & 256 & ReLU & \checkmark & \xmark & \textcolor{black}{30.80}($-$0.03) & \textcolor{black}{26.73}($+$0.00)\\
\hline
\multicolumn{1}{l|}{LTEW $+$ \cite{sitzmann2019siren}} & \checkmark & \checkmark & 256 & $\sin$ & \checkmark & \checkmark & \textcolor{black}{30.80}($-$0.03) & \textcolor{black}{26.71}($-$0.02)\\
\multicolumn{1}{l|}{LTEW} & \checkmark & \checkmark & 256 & ReLU & \checkmark & \checkmark & \textbf{\textcolor{black}{30.83}}($+$0.00) & \textcolor{black}{26.73}($+$0.00)
\end{tabular}
}
\vspace{-12pt}
\end{table*}

\noindent
In Table.~\ref{tab:Quan_Abl}, we explore other arbitrary-scale SR methods \cite{chen2021learning, itsrn} for image warping and validate our design with extensive ablation studies. Rows 1-5 show that Fourier features provide performance gain compared to concatenation \cite{chen2021learning}, transformer \cite{itsrn} or periodic activation \cite{sitzmann2019siren}. Since \cite{chen2021learning, itsrn} are designed to perform rectangular SR, we retrain them after modifying \textit{cell} in \cite{chen2021learning} (rows 2-3) and \textit{token} in \cite{itsrn} (row 5) to our shape term as Eq.~\eqref{eq:cell}. 

In rows 6-8, we remove an amplitude estimator (row 6), long skip connection (row 7), and reduce the number of estimated frequencies (row 8). We see that each component consistently enhances mPSNR of LTEW for both \textit{in-scale} and \textit{out-of-scale}. In row 9, we test LTEW with spatially-invariant cell as \cite{lee2021local}, specifically $r_\mathbf{x}=2w/W, r_\mathbf{y}=2h/H$. It causes a significant mPSNR drop for both \textit{in-scale} and \textit{out-of-scale}. From rows 10-12, we observe that both the Jacobian and Hessian shapes are significant in improving mPSNR only for \textit{in-scale}.
Inspired by \cite{DBLP:conf/iclr/XuZLDKJ21}, we hypothesize that INR performs superior interpolating unseen coordinates but relatively poorly extrapolating untrained shapes. Extrapolation for untrained shapes will be investigated in future work.

\subsection{Fourier feature space}

\begin{figure}[ht!]
\centering
\includegraphics[scale = 0.177]{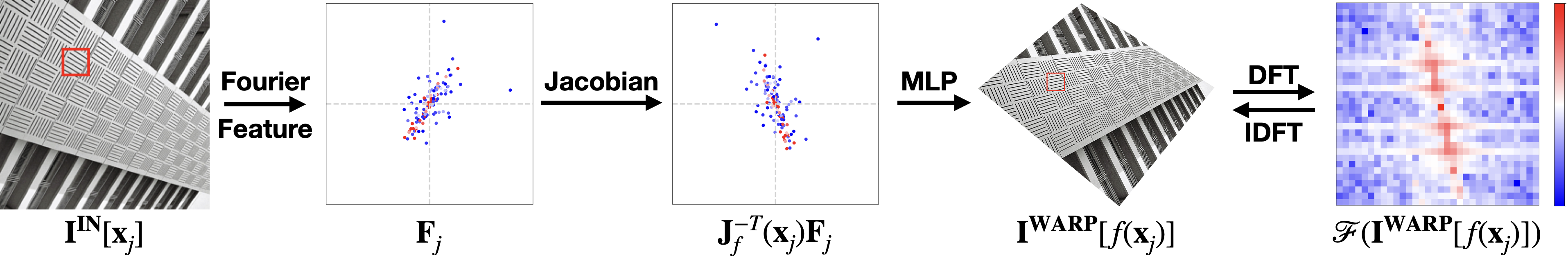}
\vspace{-6pt}
\caption{{Visualization of Fourier feature space from LTEW.}}
\vspace{-12pt}
\label{fig:fourier}
\end{figure}

In Fig.~\ref{fig:fourier}, we visualize estimated Fourier feature space from LTEW and the discrete Fourier transform (DFT) of a warped image for validation. We first scatter estimated frequencies to 2D space and set the color according to magnitude. We observe that our LTEW ($h_{\psi}$) extracts Fourier information ($\mathbf F_j$) for an input image at $\mathbf x_j$ ($\mathbf{I^{IN}}[\mathbf x_j]$). By observing pixels inside an encoder's receptive field (RF), LTEW estimates dominant frequencies and corresponding Fourier coefficients for RF-sized local patches of an input image. Before our MLP ($g_\theta$) represents $\mathbf{I^{WARP}}[f(\mathbf x_j)]$, the Fourier space ($\mathbf F_j$) is transformed by the Jacobian matrix $\mathbf J_f^{-T}(\mathbf x_j)$ and matched to a frequency response of an output image at $f(\mathbf x_j)$ ($\mathcal{F}(\mathbf{I^{WARP}}[f(\mathbf x_j)])$). We observe that LTEW utilizing the local grid in output space ($\mathbf{\delta_y}$) instead of $\mathbf{\delta_x}$ diverges during training. We hypothesize that representing $\mathbf{I^{WARP}}[f(\mathbf x_j)]$ with frequencies ($\mathbf F_j$) of an input image makes the overall training procedure unstable. This indicates that a spatially-varying Jacobian matrix $\mathbf{J}_f^{-T}(\cdot)$ for the given coordinate transformation $f$ is significant in predicting accurate frequency responses of warped images. By estimating Fourier response in latent space instead of directly applying DFT to input images, we avoid extracting undesirable frequencies due to aliasing, as discussed in \cite{lee2021local}.


\subsection{Discussion}
In Table~\ref{tab:Quan_sym}, we compare model complexity and symmetric-scale SR performance of our LTEW for both \textit{in-scale} and \textit{out-of-scale} to other warping methods: ArbSR \cite{Wang2020Learning} and SRWarp \cite{SRWarp}. Note that ArbSR \cite{Wang2020Learning} and SRWarp \cite{SRWarp} are learned to perform asymmetric-scale SR and homography transform. Following \cite{lee2021local}, we use $\max(\mathbf{s}, \mathbf{s}_{tr})$ instead of $\mathbf s$ in Eq.~\eqref{eq:twelve} for phase estimation. We see that LTEW significantly outperforms exiting warping methods for \textit{out-of-scale}, achieving competitive quality to \cite{Wang2020Learning, SRWarp} for \textit{in-scale}. A local ensemble \cite{Local_Implicit_Grid_CVPR20, chen2021learning, lee2021local} in LTEW, preventing blocky artifacts, makes the model more complex than ArbSR \cite{Wang2020Learning}. SRWarp \cite{SRWarp} blends $\times 1$, $\times 2$, and $\times 4$ features, leading to increased model complexity than LTEW, which uses only an $\times1$ feature map.

\begin{table*}[ht!]
\vspace{-12pt}
\centering
\setlength{\tabcolsep}{1.2pt}
\scriptsize{
\caption{Quantitative comparison with state-of-the-art warping methods for \underline{\textbf{symmetric-scale SR}} on B100 dataset (PSNR (dB)). RCAN \cite{zhang2018rcan} and RRDB \cite{Wang_2018_ECCV_Workshops} are used as encoders. Computation time and memory consumption are measured with an $256\times 256$-sized input for an $\times2$ task on NVIDIA RTX 3090 24GB.}
\label{tab:Quan_sym}
\vspace{-6pt}
\begin{tabular}{c
|>{\centering\arraybackslash}p{1.50cm}
|>{\centering\arraybackslash}p{1.22cm}>{\centering\arraybackslash}p{1.22cm}>{\centering\arraybackslash}p{1.22cm}
|>{\centering\arraybackslash}p{0.71cm}>{\centering\arraybackslash}p{0.71cm}>{\centering\arraybackslash}p{0.71cm}|>{\centering\arraybackslash}p{0.71cm}>{\centering\arraybackslash}p{0.71cm}
}
\multirow{2}{*}{Method} & \multirow{2}{*}{\makecell{Training\\task}} & \multirow{2}{*}{\#Params.} & \multirow{2}{*}{Runtime} & \multirow{2}{*}{Memory} & \multicolumn{3}{c|}{\textit{in-scale}} & \multicolumn{2}{c}{\textit{out-of-scale}}\\
& & & & & $\times2$ & $\times3$ & $\times4$ & $\times6$ & $\times8$ \\
\hline
Arb-RCAN \cite{Wang2020Learning} & \multirow{2}{*}{\makecell{\tiny Asymmetric\\\tiny -scale SR}}  & 16.6M & \textbf{160ms} & \textbf{1.39GB} & \textbf{32.39} & \textbf{29.32} & 27.76 & 25.74 & 24.55 \\
LTEW-RCAN (ours) & & \textbf{15.8M} & 283ms & 1.77GB & 32.36 & 29.30 & \textbf{27.78} & \textbf{26.01} & \textbf{24.95}  \\
\hline
SRWarp-RRDB \cite{SRWarp} & \multirow{2}{*}{\makecell{\tiny Homography\\\tiny transform}} & 18.3M & 328ms & 2.34GB & 32.31 & 29.27 & \textbf{27.77} & 25.33 & 24.45 \\
LTEW-RRDB (ours) & & \textbf{17.1M} & \textbf{285ms} & \textbf{1.79GB} & \textbf{32.35} & \textbf {29.29} & 27.76 & \textbf{25.98} & \textbf{24.95} \\
\end{tabular}
}
\vspace{-12pt}
\end{table*}

\section{Conclusions}

In this paper, we proposed the continuous neural representation by learning Fourier characteristics of images warped by the given coordinate transformation. Particularly, we found that shape-dependent phase estimation and long skip connection enable MLP to predict signals more accurately. We demonstrated that the LTEW-based neural function outperforms existing warping techniques for asymmetric-scale SR and homography transform. Moreover, our method effectively generalizes untrained coordinate transformations, specifically \textit{out-of-scale} and ERP perspective projection.

\small{\textbf{Acknowledgement} This work was partly supported by the National Research Foundation of Korea (NRF) grant funded by the Korea government (MSIT) (No. 2021R1A4A1028652), the DGIST R\&D Program of the Ministry of Science and ICT (No. 22-IJRP-01) and Institute of Information \& communications Technology Planning \& Evaluation (IITP) grant funded by the Korea government (MSIT) (No. IITP-2021-0-02068).}

%
%
\bibliographystyle{splncs04}

\end{document}